\documentclass[10pt,twocolumn,letterpaper]{article}

\usepackage[]{cvpr}      % To produce the REVIEW version
\usepackage{amsfonts,amssymb,amsmath,pifont}
\usepackage{multirow}
\usepackage{textcomp}
\usepackage[dvipsnames]{xcolor}

\definecolor{cvprblue}{rgb}{0.21,0.49,0.74}
\usepackage[pagebackref,breaklinks,colorlinks,citecolor=cvprblue]{hyperref}

\def\paperID{14713} % *** Enter the Paper ID here
\def\confName{CVPR}
\def\confYear{2024}

\title{GIST: Improving Parameter Efficient Fine Tuning via Knowledge Interaction}

\author{Jiacheng Ruan, Jingsheng Gao, Mingye Xie, Suncheng Xiang, Zefang Yu, Ting Liu, Yuzhuo Fu\\
Shanghai Jiao Tong University\\
{\tt\small jackchenruan@sjtu.edu.cn}
}

\begin{document}
\maketitle
\begin{abstract}
The Parameter-Efficient Fine-Tuning (PEFT) method, which adjusts or introduces fewer trainable parameters to calibrate pre-trained models on downstream tasks, has become a recent research interest. However, existing PEFT methods within the traditional fine-tiuning framework have two main shortcomings: 1) They overlook the explicit association between trainable parameters and downstream task knowledge. 2) They neglect the interaction between the intrinsic task-agnostic knowledge of pre-trained models and the task-specific knowledge in downstream tasks. To address this gap, we propose a novel fine-tuning framework, named \textbf{GIST}, in a plug-and-play manner. Specifically, our framework first introduces a trainable token, called the Gist token, when applying PEFT methods on downstream tasks. This token serves as an aggregator of the task-specific knowledge learned by the PEFT methods and forms an explicit association with downstream knowledge. Furthermore, to facilitate explicit interaction between task-agnostic and task-specific knowledge, we introduce the concept of \textbf{Knowledge Interaction} via a Bidirectional Kullback-Leibler Divergence objective. As a result, PEFT methods within our framework can make the pre-trained model understand downstream tasks more comprehensively by leveraging the knowledge interaction. Extensive experiments demonstrate the universality and scalability of our framework. Notably, on the VTAB-1K benchmark, we employ the Adapter (a prevalent PEFT method) within our GIST framework and achieve a performance boost of 2.25\%, with an increase of only 0.8K parameters (0.01‰ of ViT-B/16). The Code will be released.
\end{abstract}    
\section{Introduction}
\label{sec:intro}

\begin{figure}[t] % 图片环境，htbp参数表示浮动定位
\centering % 居中对齐
\includegraphics[width=0.48\textwidth]{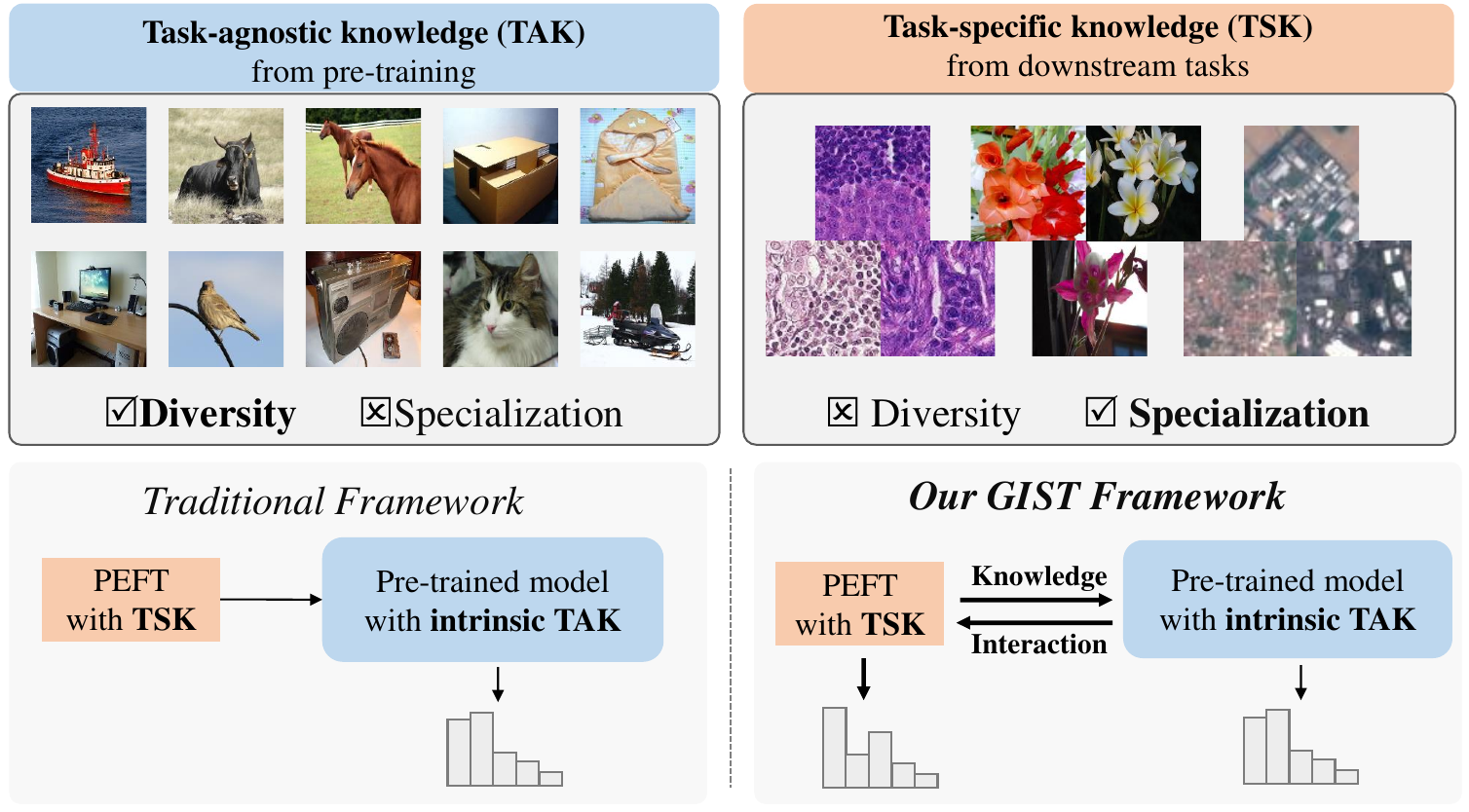} % 插入图片，width=0.5\textwidth表示图片宽度为文本宽度的一半
\caption{The issue and our motivation: Models can acquire task-agnostic knowledge (TAK) via pre-training, which is often broad and diverse but lacks specialization. Conversely, they can obtain task-specific knowledge (TSK) through fine-tuning on downstream tasks, which is typically specialized. Unlike the traditional fine-tuning framework, our goal is to establish an explicit connection between the learnable parameters and the downstream tasks, thereby comprehensively learning TSK. In addition, we introduce the concept of knowledge interaction, establishing interactions between the TAK represented by the frozen parameters and the TSK represented by the learnable parameters, thus enabling the model to better adapt to the downstream tasks.} % 图片标题
\label{figurehead}
\end{figure}

The advent of large-scale datasets and the pre-training fine-tuning paradigm has empowered pre-trained models to achieve remarkable performances \cite{tfl-book}. By leveraging task-agnostic knowledge (\textbf{\textit{TAK}}) from the pre-training phase and learning task-specific knowledge (\textbf{\textit{TSK}}) during the fine-tuning process \cite{vlm4cv-survey, tfl-survey1, tfl-survey2}, pre-trained models, particularly Transformer-based models \cite{vitsurvey, vlpsurvey}, have exhibited exemplary performance across fields such as computer vision (CV) and natural language processing (NLP). However, the burgeoning parameters in Transformer-based models have made the Full parameter fine-Tuning (FT) method less practical for downstream tasks. The FT method demands training and storing different full parameters for each task, an approach that is not only storage-unfriendly but also prone to overfitting due to the often limited data volume in downstream tasks.

To enhance fine-tuning efficiency, the research community has shown growing interest in Parameter-Efficient Fine-Tuning (PEFT) methods \cite{peftreview}. PEFT methods predominantly freeze the bulk of pre-trained model parameters, adjusting or introducing a small set of trainable parameters to assimilate TSK. However, as shown in Figure \ref{figurehead}, PEFT methods within traditional fine-tuning framework do not explicitly establish a connection between the learnable parameters and TSK, and also overlook the interaction with TAK.

To address this issue, we initiate our investigation from the perspective of TSK acquisition, based on VPT \cite{VPT}, a classic PEFT method. VPT achieves commendable fine-tuning performance by freezing the model's backbone parameters and introducing learnable prompt tokens. However, in VPT, prompt tokens are not explicitly used for the final loss calculation, which may hinder the learning of TSK by trainable parameters. Consequently, we naturally utilize prompt tokens as an additional dependency for computing the loss, observing an improvement in performance. Subsequently, to demonstrate the scalability of this discovery, we attempt to employ this loss as a plug-and-play design. We apply another PEFT method (\textit{e.g.}, Adapter) with VPT during the fine-tuning process, and find that using prompt tokens as an extra criterion also enhances performance. This discovery implies that naturally using learnable parameters as the basis for computing loss can lead to more effective downstream TSK learning.

Inspired by the above observations, we propose \textbf{GIST}, a concise and efficient framework, in a plug-and-play manner. To reduce the additional parameter burden, we directly decrease the length of prompt tokens introduced by VPT to 1, referred to as the Gist token ([GIST]), which acts as an aggregator to integrate the TSK learned by the PEFT parameters. This concept of an aggregator is derived from the pre-training stage of ViT \cite{vit}, where a Class token ([CLS]) is introduced to aggregate global information for the final loss calculation.\footnote{Note that [GIST] is introduced during downstream fine-tuning, is trainable, and aggregates TSK, while [CLS] is frozen during fine-tuning and can be considered to retain TAK.} Thus, by simply using the GIST token as an additional basis for loss calculation, it can help the learnable parameters of PEFT obtain more comprehensive TSK. Additionally, to address the lack of knowledge interaction, we introduce a Bidirectional Kullback-Leibler Divergence (BKLD) objective between [CLS] and [GIST], bridging the gap between the TAK and TSK. In short, under our GIST framework, PEFT methods can improve fine-tuning performance while introducing only an additional 0.8K parameters (0.01‰ of ViT-B/16).

We conduct extensive experiments to validate the effectiveness of our framework on 19 image classification, 5 fine-grained few-shot and 8 language understanding datasets. The results demonstrate that, when existing PEFT methods are fine-tuned within our GIST framework, they consistently deliver improved performance across a diverse range of scenarios without necessitating a substantial increase in parameter count. The contributions are as follows: 
\begin{itemize}
    \item We introduce a learnable Gist token, serving as an aggregator for task-specific knowledge acquisition. This establishes an explicit connection between learnable parameters and downstream tasks.
    \item We pioneer the concept of knowledge interaction during the fine-tuning phase by employing a Bidirectional Kullback-Leibler Divergence objective for explicit interaction between task-agnostic and task-specific knowledge. This objective more effectively utilizes the task-agnostic knowledge of the pre-trained models.
    \item We propose an innovative fine-tuning framework, dubbed as GIST. Extensive experimental results demonstrate that our framework enhances the performance of existing PEFT methods, with almost no increase in trainable parameters count.
\end{itemize}

\section{Related Works}
\label{sec:relatedworks}

\subsection{Parameter Efficient  Fine-tuning}

Parameter Efficient Fine-tuning (PEFT) methods enhance the performance of pre-trained models on downstream tasks in a power-saving and efficient manner. Essentially, PEFT techniques modify a select subset or introduce new trainable parameters during fine-tuning to assimilate TSK, thereby calibrating the model's predictions on downstream tasks. Initial explorations into PEFT were predominantly within NLP tasks, with notable methodologies including Adapter \cite{nlpadapter}, Prompt \cite{nlpprompt}, Prefix \cite{nlpprefix}, and LoRA \cite{nlplora}, etc. Subsequently, VPT \cite{VPT} migrates the Prompt technique from NLP to CV, demonstrating the potential of PEFT in visual tasks. For instance, VPT achieves impressive results by fine-tuning with only 0.1\% of the total model parameters. AdaptFormer \cite{adaptformer} introduces Adapter in parallel into the ViT's FFN layer, achieving performance comparable to the FT method in image recognition and video understanding tasks. SSF \cite{ssf} adjusts the model's features by scaling and shifting, achieving superior results in image recognition. This success catalyzes further research into PEFT for the CV tasks, with methodologies like Convpass \cite{convpass}, FacT \cite{fact}, and ReAdapter \cite{readapter} further advancing the state-of-the-art. However, under the traditional fine-tuning framework, existing PEFT methods do not fully realize their potential because they overlook the explicit connection with TSK and the knowledge interaction with TAK. Therefore, with hardly any increase in parameters, we propose the GIST fine-tuning framework to establish explicit connections and interactions, thereby maximizing the capabilities of existing PEFT methods.

\subsection{Self-Knowledge Distillation}

A concept resonating with our proposed framework is self-knowledge distillation. Knowledge distillation paradigms \cite{kd} focus on enhancing the performance of student models by assimilating  knowledge from a larger teacher model. In contrast, self-knowledge distillation posits the student model as its own teacher. This is achieved by deriving soft labels through specially crafted branches or distinct distributions, subsequently computing the distillation loss against its own predictions. For instance, in the BYOT approach \cite{BYOT}, the deepest classifier is regarded as the teacher, and it imparts its knowledge to shallower networks. CS-KD \cite{CSKD} uses two different samples from the same category to normalize the consistency between two different views of the predicted distribution. USKD \cite{uskd} utilizes the student model's logits as soft target labels and employs the ranking of intermediate features along with Zipf's law to generate soft non-target labels. Subsequently, USKD performs knowledge distillation using both soft labels from target and non-target classes, making it an advanced approach. In our work, we extrapolate the concept of knowledge distillation. By introducing a learnable token to derive soft labels and employing the BKLD loss as the metric between these soft labels and the model's predictions, our fine-tuning framework aims to augment the efficacy of extant PEFT techniques with negligible parameter overhead.

\section{Methods}
This section delineates our GIST framework. Initially, in Section \ref{sec.preexp}, we reassess the PEFT methods from a knowledge perspective, offering a potential application at the framework level. Subsequently, Section \ref{sec.gistframework} delves into our GIST framework, elucidating the integration of the Gist token within the Transformer architecture, and the employment of the Bidirectional Kullback-Leibler Divergence (BKLD) loss for knowledge interaction.

\subsection{Rethinking PEFT via knowledge acquisition}
\label{sec.preexp}

\begin{figure}[t] % 图片环境，htbp参数表示浮动定位
\centering % 居中对齐
\includegraphics[width=0.45\textwidth]{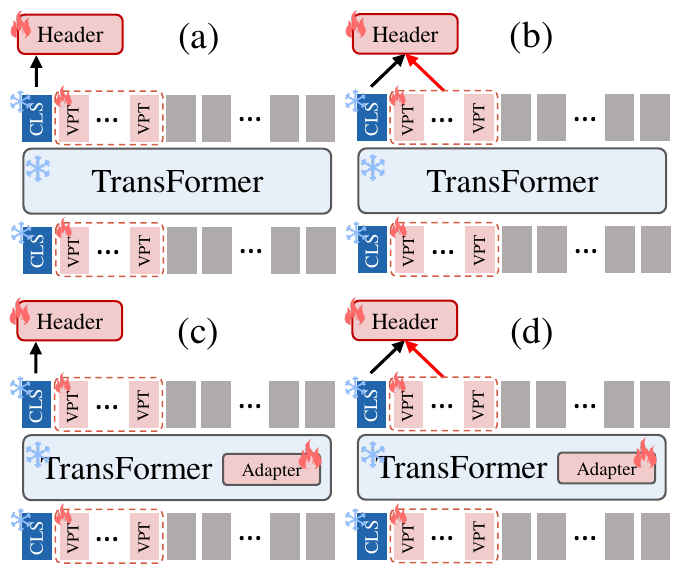} % 插入图片，width=0.5\textwidth表示图片宽度为文本宽度的一半
\caption{The different fine-tuning structures in Table \ref{tablepreexp}. (a) original VPT in \cite{VPT}. (b) on the top of (a), additionally using VPT prompt tokens as the basis for calculating loss. (c) combining two classic PEFT methods (VPT and Adapter) for fine-tuning. (d) on the top of (c), additionally using VPT prompt tokens as the basis for calculating loss.} % 图片标题
\label{figurepreexp}
\end{figure}

\begin{table}[t]
\centering
\footnotesize
\begin{tabular}{c|cc|cc}
\hline
\textbf{Tag} & \textbf{Method} & \textbf{Loss} & \textbf{Params. (M)} & \textbf{Mean} \\ \hline
(a) &VPT & \( \mathcal{L}_{ce} \) & 0.05 & 62.41 \\ 
(b) &VPT & \( \mathcal{L}_{ce} + \mathcal{L}_{vpt} \) & 0.05 & 62.91 \\ 
\hline
- &Adapter & \( \mathcal{L}_{ce} \)  & 0.13 & 71.46 \\ 
(c) &Adapter + VPT & \( \mathcal{L}_{ce} \) & 0.15 & 71.70 \\ 
(d) &Adapter + VPT & \( \mathcal{L}_{ce} + \mathcal{L}_{vpt} \) & 0.15 & 72.19 \\ 
\hline
\end{tabular}
\caption{Top-1 average accuracy on VTAB-1K. We are progressively experimenting with various structural combinations to enhance performance on downstream tasks. The hidden dimension of the Adapter is 4, and the prompt tokens' length introduced by VPT is 20. It is better viewed in conjunction with Figure \ref{figurepreexp}.}
\label{tablepreexp}
\end{table}

\begin{figure*}[t] % 图片环境，htbp参数表示浮动定位
\centering % 居中对齐
\includegraphics[width=0.9\textwidth]{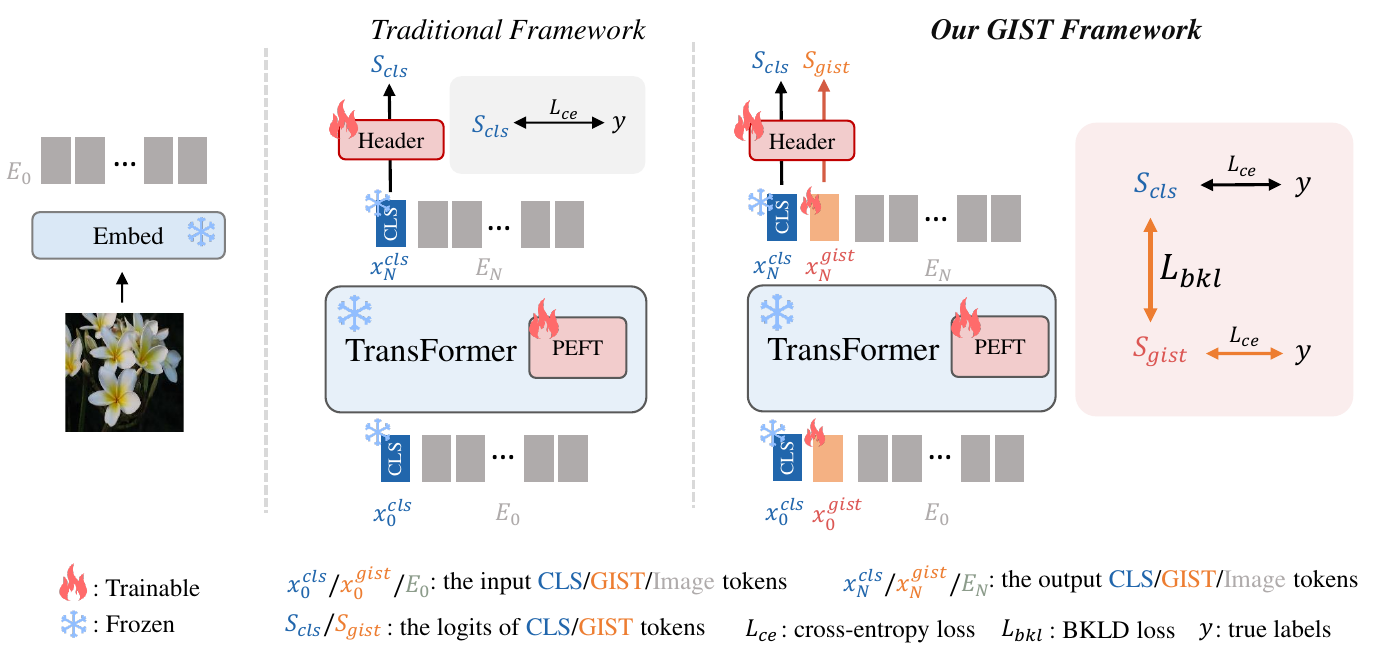} % 插入图片，width=0.5\textwidth表示图片宽度为文本宽度的一半
\caption{Overview of our GIST fine-tuning framework. Unlike the traditional fine-tuning framework, we introduce a learnable token of length one, called the Gist token, which collaboratively learns task-specific knowledge with the PEFT method on downstream tasks. Subsequently, we introduce a Bidirectional Kullback-Leibler Divergence loss to facilitate knowledge interaction.} % 图片标题
\label{figuremain}
\end{figure*}

In this section, we first explore existing PEFT methods from the perspective of knowledge acquisition. Experiments are conducted on the VTAB-1K benchmark, with settings identical to those in Section \ref{sec.implements}. Initially, as shown in Figure \ref{figurepreexp}(a), we start our exploration with a classic PEFT method, VPT-shallow \cite{VPT}. We fix the length of the learnable prompt tokens introduced by VPT to 20, achieving an accuracy of 62.41\% in fine-tuning (Table \ref{tablepreexp}(a)). However, in the original VPT, only the Class token is utilized to calculate cross-entropy loss with true labels, and the learnable prompt tokens are not directly involved in the loss computation. We believe this form is suboptimal for fine-tuning. Therefore, as shown in Figure \ref{figurepreexp}(b) and Equation \ref{eq_rvpt}, we naturally attempt to incorporate the prompt tokens for calculating the cross-entropy loss with the true labels. This simple modification results in a 0.5\% increase in fine-tuning performance (Table \ref{tablepreexp}(b)). A possible reason is that explicitly including learnable parameters in the loss calculation can lead to more comprehensive task-specific knowledge (\textbf{\textit{TSK}}) acquisition.

\begin{equation}
\begin{aligned}
&\mathcal{L} = \mathcal{L}_{ce}(S_{cls},y) + \mathcal{L}_{vpt} \\
&\mathcal{L}_{vpt} = \mathcal{L}_{ce}(S_{vpt},y)
\end{aligned}
\label{eq_rvpt}
\end{equation} where $\mathcal{L}_{ce}$ denotes the cross-entropy loss, and $y$ represents true labels. $S_{cls}$ and $S_{vpt}$ represent the logits obtained from the Class token and VPT prompt tokens after passing through the linear classification head, respectively.

Subsequently, we investigate the feasibility of integrating the loss calculation approach derived from VPT with other PEFT methods. As shown in Figures \ref{figurepreexp}(c, d), VPT is implemented alongside the Adapter for the fine-tuning process. Performance is evaluated and compared before and after the incorporation of the additional loss $\mathcal{L}_{vpt}$. The results indicate that this supplementary loss, detailed in Tables \ref{tablepreexp}(c, d), further enhances the Adapter's fine-tuning efficacy.

Therefore, we pose a question: \textit{Can this method serve as a free lunch-style framework to enhance the fine-tuning performance of existing PEFT methods?} The answer is affirmative. In the next section , we introduce our GIST fine-tuning framework, which can enhance the performance of PEFT methods in a plug-and-play manner without adding extra burden.

\subsection{GIST Framework}
\label{sec.gistframework}

As discussed in Section \ref{sec.preexp}, incorporating VPT's prompt tokens for additional loss calculation can enhance fine-tuning performance. However, this approach also introduces an increased parameter burden. Additionally, relying solely on $\mathcal{L}_{vpt}$ as an extra loss component does not fully utilize the task-agnostic knowledge (\textbf{\textit{TAK}}) from the pre-training phase. Therefore, as shown in Figure \ref{figuremain}, our GIST framework introduces a special token called the Gist token, which is only 1 in length and designed to be an aggregator for learning TSK. Furthermore, we introduce the BKLD loss for knowledge interaction, thereby maximizing the potential of PEFT methods for more effective fine-tuning.

\paragraph{Gist token ([GIST])}

% Referencing Figure \ref{figuremain}  (a), 
For a TransFormer model\footnote{In this section, for the sake of simplicity in expression, we omit the processing procedure of the PEFT method.}, the input data undergoes an embedding transformation, yielding a sequence $x\in\mathbb{R}^{L\times D}$. Afterwards, the Class token $x_0^{cls}\in\mathbb{R}^{1\times D}$ is concatenated with this sequence, augmented by a positional embedding $P\in\mathbb{R}^{(L+1)\times D}$, resulting in the input sequence $X_0\in\mathbb{R}^{(L+1)\times D}$, as formulated in Equation \ref{eq4}.

\begin{equation}
X_0 = [x_0^{cls};x]+P
\label{eq4}
\end{equation}where $[\cdot;\cdot]$ represents the concatenation operation. Subsequent processing of $X_0$ ensues through a series of Transformer layers, as depicted in Equation \ref{eq5}.

\begin{equation}
\begin{aligned}
&X^{\prime}_l=\textup{MHSA}(\textup{LN}(X_{l-1}))+X_{l-1}\\
&X_l=\textup{FFN}(\textup{LN}(X^{\prime}_l))+X^{\prime}_l, l=1,2,...,N
\end{aligned}
\label{eq5}
\end{equation} where $\textup{MHSA}$ stands for multi-head self-attention block, $\textup{FFN}$ represents the feed-forward network, and $\textup{LN}$ stands for LayerNorm \cite{layernorm}. After applying all Transformer layers, we can derive $x_N^{cls}$ from $X_N$. The logits $S_{cls}$ can then be obtained using the linear classification head ($\textup{HEAD}$), as shown in Equation \ref{eq6}.

\begin{equation}
S_{cls} = \textup{HEAD}(x_N^{cls})
\label{eq6}
\end{equation}

Ultimately, the cross entropy loss function computes the loss between $S_{cls}$ and the true labels. Notably, during the fine-tuning phase, $x_0^{cls}$ is frozen, preserving the model's TAK from the pre-training phase. Different from the traditional fine-tuning framework as shown in Figure \ref{figuremain}, our framework introduces an additional learnable token ([GIST]), denoted as $x_0^{gist}\in\mathbb{R}^{1\times D}$, to aggregate the TSK learned by the PEFT method during fine-tuning. Thus, on the basis of Equation \ref{eq4}, we concatenate [GIST] to obtain our input sequence $X_0\in\mathbb{R}^{(L+2)\times D}$, as Equation \ref{eq7}.

\begin{equation}
X_0 = [[x_0^{cls};x]+P;x_0^{gist}]
\label{eq7}
\end{equation}

After processing the input sequence through all Transformer layers, $x_N^{cls}$ and $x_N^{gist}$ are derived from $X_N$. We subsequently send both $x_N^{cls}$ and $x_N^{gist}$ through the linear classification head, resulting in $S_{cls}$ and $S_{gist}$, respectively. The loss $\mathcal{L}_{cls}$ is computed by contrasting $S_{cls}$ with the truth labels. Similarly, the loss $\mathcal{L}_{gist}$ is determined by comparing $S_{gist}$ with the true labels. These two loss terms can be expressed as follows:

\begin{equation}
\begin{aligned}
\mathcal{L}_{cls} = \mathcal{L}_{ce}(S_{cls}, y) \\
\mathcal{L}_{gist} = \mathcal{L}_{ce}(S_{gist}, y)
\label{eq8}
\end{aligned}
\end{equation}where $\mathcal{L}_{ce}$ is the cross entropy loss, $y$ is the true labels.

\paragraph{Bidirectional Kullback-Leibler Divergence (BKLD) Loss}

Only utilizing  $\mathcal{L}_{cls}$ and $\mathcal{L}_{gist}$ does not facilitate explicit interaction between the TAK represented by $S_{cls}$ and the TSK represented by $S_{gist}$. Therefore, we introduce the BKLD loss function, as shown in Equation \ref{eq9}.

\begin{equation}
\begin{aligned}
\mathcal{L}_{bkl} &= \mathcal{L}_{fkl} + \mathcal{L}_{rkl} \\
&= \textup{KL}(S_{cls}||S_{gist};T)+\textup{KL}(S_{gist}||S_{cls};T)
\label{eq9}
\end{aligned}
\end{equation} where $\mathcal{L}_{bkl}$ represents our BKLD loss. $\mathcal{L}_{fkl}$ is the forward KLD loss. $\mathcal{L}_{rkl}$ is the reverse KLD loss. $\textup{KL}(\cdot||\cdot;T)$ means computing the KL divergence between two distributions with a temperature $T$. The parameter $T$, is introduced to soften the outputs before they are processed through softmax, adjusting the sharpness of the distribution. Higher values of $T$ produce softer probabilities \cite{kdsurvey}.

For most knowledge distillation methods, the forward KLD is generally utilized as the loss function. It can be represented as $\mathcal{L}_{fkl}= \textup{KL}(p||q;T)$, where $p$ and $q$ represent two different distributions.  With $p$ taken as the reference, $\mathcal{L}_{fkl}$ quantifies how much the distribution $q$ diverges from $p$. Conversely, the reverse KLD, denoted as $\mathcal{L}_{rkl}=\textup{KL}(q||p;T)$, uses $q$ as the reference and measures the divergence of distribution $p$ from $q$. In this paper, we leverage both forward and reverse KLD as loss functions to facilitate explicit interaction between TAK and TSK. On one hand, we employ the forward KLD loss to enhance the learning of TSK, guided by TAK. On the other hand, by utilizing the reverse KLD loss, we ensure that the pre-trained model is more effectively tailored to downstream tasks, following the directives of TSK.

\paragraph{Overall Loss} The overall loss function is derived by amalgamating $\mathcal{L}_{cls}$, $\mathcal{L}_{gist}$, and $\mathcal{L}_{bkl}$, as depicted in Equation \ref{eq10}. This loss function guides the model during the fine-tuning phase, allowing [GIST] to co-learn with other PEFT parameters and aggregate TSK, while fully leveraging TAK to ensure an explicit interaction between the two types of knowledge.

\begin{equation}
\mathcal{L}_{all} = \mathcal{L}_{cls} + \mu \mathcal{L}_{gist} + \lambda \mathcal{L}_{bkl}
\label{eq10}
\end{equation} where $ \mu$ and $\lambda$ is the hyperparameter that controls the trade-off among the three loss terms. It is noteworthy that the aforementioned use of $S_{gist}$ is limited only to the training process. For the inference process, we still solely rely on $S_{cls}$ as the exclusive basis for prediction.
\begin{table*}[t]
\centering
\footnotesize
\setlength{\tabcolsep}{2pt}
\begin{tabular}{c|ccccccc|cccc|cccccccc|ccc}
\hline
\multicolumn{1}{l|}{} & \multicolumn{7}{c|}{Natural}                                      & \multicolumn{4}{c|}{Specialized}                  & \multicolumn{8}{c|}{Structured}                                                                                     & \multicolumn{1}{l}{} & \multicolumn{1}{l}{} \\ \hline
Method             & \rotatebox{90}{CIFAR-100} & \rotatebox{90}{Caltech101} & \rotatebox{90}{DTD}  & \rotatebox{90}{Flowers102} & \rotatebox{90}{Pets} & \rotatebox{90}{SVHN} & \rotatebox{90}{Sun397} & \rotatebox{90}{Patch Camelyon} & \rotatebox{90}{EuroSAT} & \rotatebox{90}{Resisc45} & \rotatebox{90}{Retinopathy} & \rotatebox{90}{Clevr/count} & \rotatebox{90}{Clevr/distance} & \rotatebox{90}{DMLab} & \rotatebox{90}{KITTI/distance} & \rotatebox{90}{dSprites/loc} & \rotatebox{90}{dSprites/ori} & \rotatebox{90}{SmallNORB/azi} & \rotatebox{90}{SmallNORB/ele} & \rotatebox{90}{Mean}        &{$\Delta$}          & \rotatebox{90}{Params. (M)}          \\ \hline
FT                    & 68.9      & 87.7       & 64.3 & 97.2       & 86.9 & 87.4 & 38.8   & 79.7           & 95.7    & 84.2     & 73.9        & 56.3        & 58.6           & 41.7  & 65.5           & 57.5         & 46.7         & 25.7          & 29.1          & 65.57        &   -     & 85.84                \\
LP        & 63.4      & 85.0         & 63.2 & 97.0         & 86.3 & 36.6 & 51.0     & 78.5           & 87.5    & 68.6     & 74.0          & 34.3        & 30.6           & 33.2  & 55.4           & 12.5         & 20.0           & 9.6           & 19.2          & 52.94       &    -     & 0.04                 \\ \hline
Adapter               & 70.2      & 92.6       & 74.6 & 99.4       & 91.2 & 80.4 & 51.4   & 84.1           & 96.3    & \textbf{88.0}       & 75.6        & \textbf{84.2}        & 59.6           & 53.2  & 76.3           & 60.7         & 51.9         & 27.8          & 40.2          & 71.46       &  \multirow{2}{*}{\textbf{2.25\textuparrow}}      & 0.13                 \\ 
\textbf{Adapter\textasteriskcentered}          &\textbf{74.5}     & 92.3       & \textbf{76.9} & \textbf{99.5}       & \textbf{92.3} & 85.7 & 54.6   & \textbf{88.2}           & \textbf{96.5}    & 87.9     & \textbf{77.4}        & 83.6        & 61.2           & 54.0    & \textbf{81.2}           & 72.3         & 52.1         & 29.3          & 41.0            & 73.71      &      & 0.13                 \\ \hline
VPT                   & 60.5      & 90.6       & 70.6 & 99.1       & 89.3 & 50.1 & 50.8   & 82.2           & 93.8    & 82.5     & 74.9        & 50.6        & 58.9           & 41.0    & 68.1           & 39.0           & 32.4         & 22.3          & 29.1          & 62.41         & \multirow{2}{*}{\textbf{1.22\textuparrow}}     & 0.05                 \\
\textbf{VPT\textasteriskcentered}              & 64.6      & 90.9       & 72.3 & 99.3       & 90.4 & 56.4 & 52.6   & 82.8           & 93.9    & 83.6     & 75.1        & 49.0          & 60.5           & 41.1  & 66.9           & 43.0           & 34.8         & 22.7          & 29.1          & 63.63        &        & 0.05                 \\ \hline
SSF                   & 69.0        & 92.6       & 75.1 & 99.4       & 91.8 & 90.2 & 52.9   & 87.4           & 95.9    & 87.4     & 75.5        & 75.9        & 62.3           & 53.3  & 80.6           & 77.3         & \textbf{54.9}         & 29.5          & 37.9          & 73.10         &    \multirow{2}{*}{\textbf{0.91\textuparrow}}    & 0.24                 \\
\textbf{SSF\textasteriskcentered}              & 74.2      & \textbf{93.1}       & 74.4 & \textbf{99.5}       & 91.8 & 91.2 & 53.7   & 87.5           & 96.1    & 87.3     & 76.2        & 79.1        & 61.6           & \textbf{54.5}  & \textbf{81.2}          & 81.7         & 53.9         & 30.9          & 38.2          & 74.01      &         & 0.24                 \\ \hline
FacT                  & 70.6      & 90.6       & 70.8 & 99.1       & 90.7 & 88.6 & 54.1   & 84.8           & 96.2    & 84.5     & 75.7        & 82.6        & 68.2           & 49.8  & 80.7           & 80.8         & 47.4         & 33.2          & 43.0            & 73.23        &   \multirow{2}{*}{\textbf{0.32\textuparrow}}     & 0.11                 \\
\textbf{FacT\textasteriskcentered}             & 71.0        & 91.8       & 70.2 & 99.0         & 90.8 & 89.3 & 54.1   & 85.7           & 95.5    & 84.3     & 75.6        & 83.2        & 69.2           & 50.3  & 80.2           & 81.4         & 47.6         & 35.2          & \textbf{43.1}          & 73.55        &        & 0.11                 \\ \hline
ReAdapter             & 72.4      & 91.6       & 71.0   & 99.2       & 91.4 & 90.7 & 55.1   & 85.3           & 95.9    & 84.6     & 75.9        & 82.3        & 68.0             & 50.4  & 79.9           & 80.4         & 49.2         & \textbf{38.6}          & 41.0            & 73.83           &   \multirow{2}{*}{\textbf{0.43\textuparrow}}  & 0.22                 \\
\textbf{ReAdapter\textasteriskcentered}        & 73.4      & 92.7       & 71.5 & 99.2       & 91.5 & \textbf{91.4} & \textbf{55.4}   & 84.9           & 96.3    & 85.2     & 75.6        & 82.6        & \textbf{70.2}           & 51.2  & 80.9           & \textbf{82.0}           & 47.3         & 36.9          & 42.8          & \textbf{74.26}         &       & 0.22                 \\ \hline
\end{tabular}
\caption{\textbf{The comparative results on VTAB-1K}. The symbol \textbf{\textasteriskcentered} indicates employing the PEFT method within our GIST framework. FT represents the full parameter fine-tuning method, and LP stands for the Linear Probing method. Params. stands for trainable parameters.}
\label{tablecv}
\end{table*}

\begin{figure*}[t] % 图片环境，htbp参数表示浮动定位
\centering % 居中对齐
\includegraphics[width=0.9\textwidth]{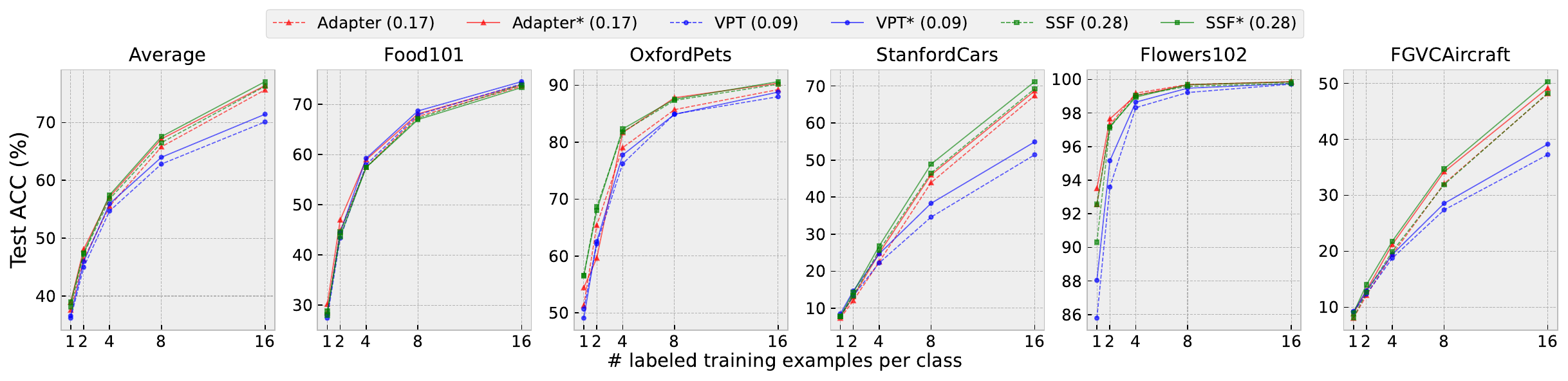} % 插入图片，width=0.5\textwidth表示图片宽度为文本宽度的一半
\caption{\textbf{Top-1 accuracy of few-shot learning on FGVC datasets}. The trainable parameters (M) is shown in parentheses.} % 图片标题
\label{figurefs}
\end{figure*}

\section{Experiments}
\subsection{Datasets and metrics}
\textbf{Image classification tasks} We utilize the VTAB-1K benchmark \cite{vtab1k} to validate our GIST framework for image classification tasks. Specifically, VTAB-1K includes 19 different datasets, which can be categorized into three groups: Natural, Specialized, and Structured. Each dataset consists of 1,000 samples for training, with an average of 20,000 samples for testing, making it a highly challenging benchmark. Following previous works \cite{ssf}, for each dataset, we report the Top-1 accuracy on the test set. For the entire benchmark, we present the arithmetic mean of the Top-1 accuracy.

\textbf{Fine-grained few-shot tasks} In a few-shot setting, we validate the performance of our framework in the low-data regime using Food-101 \cite{FGVC-food101}, OxfordPets \cite{FGVC-pets}, Stanford Cars \cite{FGVC-cars}, Oxford-Flowers102 \cite{FGVC-flowers102}, and FGVC-Aircraft \cite{FGVC-aircraft} datasets. Similar to previous work \cite{Neuralpromptsearch, fact}, we conduct validation under \{1, 2, 4, 8, 16\}-shot settings and report the Top-1 accuracy on the test set.

\textbf{Language understanding tasks} To validate the universality of our framework, we also conduct verification for the PEFT methods in NLP. GLUE benchmark \cite{glue} is utilized to verify the effectiveness of GIST framework. Specifically, we  train and test on a total of 8 tasks: MNLI, QQP, QNLI, SST-2, STS-B, MRPC, RTE, and CoLA. Following previous works \cite{attempt}, we use Pearson Correlation for STS-B and accuracy for other tasks as metrics.

\subsection{Implementation details}
\label{sec.implements}

For the VTAB-1K benchmark and FGVC datasets, we employ the ViT-B/16 \cite{vit} model, pre-trained on the ImageNet-21K dataset \cite{imagenet}, as the backbone. In terms of training configurations, we follow the work of predecessors \cite{ssf, fact, readapter}, to ensure fairness and reproducibility. Turning to the GLUE benchmark, we harness the T5-base \cite{T5} model as the backbone. Similar to the setting of the previous work \cite{attempt} by configuring a batch size of 32, imposing a maximum token length of 256, setting the learning rate to 3e-4, and conducting training for 20 epochs on each task.

Regarding our GIST framework, to avoid redundancy brought about by further hyperparameter adjustment, we fix temperature $T$ at 3, $\mu$ to 0.5, and only allow $\lambda$ to be searched from $\{0.25, 0.5, 0.75\}$. Pytorch \cite{pytorch} and Transformers \cite{huggingface} are utilized to implement experiments on NVIDIA RTX 3090 GPUs, and more detailed settings are in the Appendix.

\subsection{Main results}

\paragraph{Comparative Results on VTAB-1K}
We have thoroughly validated GIST framework on the benchmark for visual tasks, and the experimental results are shown in Table \ref{tablecv}. For the three types of PEFT methods in \cite{VisualTuning}, namely Adapter Tuning (Adapter and ReAdapter \cite{readapter}), Prompt Tuning (VPT \cite{VPT}), and Parameter Tuning (SSF \cite{ssf} and FacT \cite{fact}), we apply these methods within our framework for fine-tuning on downstream tasks. This further improves the performance of the existing PEFT methods with an average increase of 1.03\%, without significantly increasing the number of parameters. In the best case, our GIST can improve Adapter's performance by 2.25\%, and in the worst case, it can still enhance FacT's performance by 0.32\%. The results indicate that our framework facilitates a more comprehensive knowledge interaction and enhances the performance of PEFT methods by leveraging task-agnostic knowledge.

\paragraph{Comparative Results on FGVC} We conduct thorough validation in a few-shot scenario. The PEFT methods used are Adapter, VPT, and SSF, which are fine-tuned under both traditional frameworks and the GIST framework, with results shown in Figure \ref{figurefs}. Overall, even in the low-regime few-shot scenario, fine-tuning different types of PEFT methods under the GIST framework can improve performance without significantly increasing the number of trainable parameters.

\paragraph{Comparative Results on GLUE}
We conduct validation on the GLUE benchmark for NLP tasks, and the results are shown in Table \ref{tablenlp}. When applying the FT method, 220M parameters are used to achieve a performance of 84.95\%. When applying Adapter \cite{nlpadapter} within the traditional fine-tuning framework, 1.9M parameters are needed, but the performance is still 0.5\% lower than that of the FT method. Notably, when utilizing Adapter within our fine-tuning framework, the performance improves by 1.45\%, even exceeding the FT method by 0.95\%. 
\begin{table}[t]
\centering
\footnotesize
\setlength{\tabcolsep}{2pt}
\begin{tabular}{c|cccccccc|cc}
\hline
Method       & \rotatebox{90}{MNLI} & \rotatebox{90}{QQP}  & \rotatebox{90}{QNLI} & \rotatebox{90}{SST-2}   & \rotatebox{90}{STS-B} & \rotatebox{90}{MRPC} & \rotatebox{90}{RTE}  & \rotatebox{90}{CoLA} & \rotatebox{90}{Mean}  & \rotatebox{90}{Params. (M)}  \\ \hline
FT  & 86.8 & \textbf{91.6} & 93.0 & \textbf{94.6}    & 89.7  & \textbf{90.2} & 71.9 & 61.8 & 84.95    &220                   \\
Adapter      & 86.5 & 90.2 & \textbf{93.2} & 93.8    & 90.7  & 85.3 & 71.9 & 64.0 & 84.45  &1.9                    \\ 
\textbf{Adapter\textasteriskcentered} & \textbf{86.9} & 90.6 & \textbf{93.2} & 94.0 & \textbf{90.8}  & 88.7 & \textbf{77.7} & \textbf{65.3} &\textbf{85.90} &1.9  \\ \hline
\end{tabular}
\caption{\textbf{The comparative results on the GLUE benckmark}. We use Pearson Correlation for STS-B, and accuracy for other tasks as metrics. The symbol \textbf{\textasteriskcentered} indicates employing the PEFT method within our GIST framework.}
\label{tablenlp}
\end{table}

\subsection{Ablation studies}

We conduct extensive ablation experiments on the VTAB-1K benchmark. Unless otherwise specified, we employ the ViT-B/16 model, pre-trained on the ImageNet-21K dataset, as the backbone, and use Adapter as the PEFT method. Furthermore, the symbol \textbf{\textasteriskcentered} indicates employing the PEFT method within our GIST framework, and we display the arithmetic mean of the Top-1 accuracy. More detailed results can be found in the Appendix.

\paragraph{The impact of $\lambda$}

\begin{table}[t]
 \begin{minipage}[c]{0.48\linewidth}
\centering
\footnotesize
\begin{tabular}{cc}
\hline
$\lambda$        & Mean      \\ \hline
 - &71.46 \\
0.25 & 73.31  \\
0.5 & 73.18  \\
0.75 & \textbf{73.44}   \\ \hline
\end{tabular}
\caption{Ablation studies for different $\lambda$.}
\label{tablelambda}
  \end{minipage}
  \hfill
\begin{minipage}[c]{0.48\linewidth}
\centering
\footnotesize
    \setlength{\tabcolsep}{2pt}
\begin{tabular}{ccc}
\hline
token len.        & Mean  & Params. (M)    \\ \hline
 1 &73.71 & 0.13\\
10 & 73.42  & 0.14\\ 
50 & 71.96   & 0.16\\
100 & 71.21   & 0.21 \\ \hline
\end{tabular}
\caption{Ablation studies for different Gist token length.}
\label{tablelen}
  \end{minipage}

\end{table}

In our GIST framework, we only search for $\lambda$ from the set $\{0.25, 0.5, 0.75\}$ to control the interaction strength between task-agnostic and task-specific knowledge. Therefore, we first conduct ablation experiments for different interaction strengths, and the results are shown in Table \ref{tablelambda}. The results indicate that regardless of the interaction strength, our fine-tuning framework can further enhance the performance of existing methods. Even in the worst case with $\lambda=0.5$, there's still an improvement of nearly 2\%.

\paragraph{The impact of token length}

\begin{table}[t]
 \begin{minipage}[c]{0.48\linewidth}
    \centering
    \footnotesize
    \setlength{\tabcolsep}{2pt}
\begin{tabular}{ccc|c}
 \hline
$\mathcal{L}_{cls}$   & $\mathcal{L}_{gist}$   &  $\mathcal{L}_{bkl}$  & Mean  \\ \hline
\checkmark & & & 71.46 \\
\checkmark & \checkmark & & 72.71 \\
\checkmark & & \checkmark &73.29 \\
\checkmark & \checkmark & \checkmark &73.71 \\
\hline
\end{tabular}
\caption{Ablations on our loss function.}
\label{tableourloss}
  \end{minipage}
\hfill
 \begin{minipage}[c]{0.48\linewidth}
    \centering
    \footnotesize
    \setlength{\tabcolsep}{2pt}
\begin{tabular}{lc}
\hline
\multicolumn{1}{c}{Loss function} & Mean  \\  \hline
$\mathcal{L}_{cls}$              & 71.46 \\
$\mathcal{L}_{cls}$+$\mathcal{L}_{gist}$+$\mathcal{L}_{mse}$               & 73.03 \\
$\mathcal{L}_{cls}$+$\mathcal{L}_{gist}$+$\mathcal{L}_{cos}$ & 72.88 \\
$\mathcal{L}_{cls}$+$\mathcal{L}_{gist}$+$\mathcal{L}_{bkl}$              & \textbf{73.71} \\  \hline
\end{tabular}
\caption{Results on different loss functions for knowledge interaction.}
\label{tabledifferentloss}
  \end{minipage}
\end{table}

As depicted in Table \ref{tablelen}, we assess the GIST framework's performance across varying Gist token lengths. The result suggests a clear trend: as token length increases, the efficacy of our fine-tuning framework diminishes. Similar to the Class token's role during the pre-training phase, where it accumulates task-agnostic knowledge from the diverse training data, our Gist token's purpose is to aggregate downstream task knowledge during the fine-tuning phase. However, it is note that the length of the Class token is is fixed at one. Increasing the Gist token's length may cause disproportionate knowledge interaction, leading to a decline in performance.

\paragraph{The impact of loss function}

In this study, we employ loss functions that extend beyond traditional classification loss, encompassing two components: $\mathcal{L}_{gist}$ and $\mathcal{L}_{bkl}$. To evaluate the individual contributions of these components, we execute ablation studies, the results are presented in Table \ref{tableourloss}. Evidently, the efficacy of the GIST framework diminishes with a reduction in the number of loss terms. We first demonstrate the importance of establishing a direct connection between learnable parameters and task-specific knowledge during the fine-tuning process. When we introduce $\mathcal{L}_{gist}$ into the basic loss function $\mathcal{L}_{cls}$, the accuracy improved by 1.25\%. Alternatively, by adding $\mathcal{L}_{bkl}$ to $\mathcal{L}_{cls}$, it achieves a performance gain of 1.83\%, underscoring the effectiveness of knowledge interaction during downstream fine-tuning. Finally, when we introduce both types of losses simultaneously, the overall performance improves by 2.25\%. This not only proves the compatibility of these two loss functions but also indicates that more comprehensive downstream knowledge acquisition can enhance the effects of knowledge interaction.

Furthermore, we assess the performance of our framework by substituting the BKLD loss with the Mean Squared Error loss $\mathcal{L}_{mse}$ and the Cosine Similarity loss $\mathcal{L}_{cos}$. The comparative results are depicted in Table \ref{tabledifferentloss}. Intriguingly, within the confines of our GIST framework, replacing our BKLD loss by common loss functions for knowledge interaction still yields a performance enhancement ranging from 1\% to 2\%. This attests to the scalability of our fine-tuning framework. Namely, when more advanced loss functions are proposed in subsequent research, our GIST framework can also be utilized directly to enhance the performance of the existing PEFT methods.

\paragraph{The impact of different networks}

\begin{table}[t]
  \centering
\setlength{\tabcolsep}{1pt}
  \begin{minipage}[b]{0.45\linewidth}
    \centering
    \footnotesize
    \begin{tabular}{c|cc}
      \hline
      Method  &Params  & Mean \\ \hline
      S+Adapter & 0.07 & 71.39 \\
\textbf{S+Adapter\textasteriskcentered} & 0.07 & \textbf{72.47} \\  \hline
      L+Adapter & 0.30 & 71.81 \\
\textbf{L+Adapter\textasteriskcentered} & 0.30 & \textbf{73.89} \\
      \hline
    \end{tabular}
    \caption{Results on ViT-S/16 (S) and ViT-L/16 (L).}
    \label{tablevit}
  \end{minipage}
  \hfill
  \begin{minipage}[b]{0.45\linewidth}
  \footnotesize
    \centering
    \begin{tabular}{c|cc}
      \hline
      Method  &Params  & Mean \\ \hline
      FT & 86.7 & 72.46 \\
      Linear probing & 0.1 & 58.19 \\  \hline
      Adapter & 0.21 & 73.19 \\
\textbf{Adapter\textasteriskcentered} & 0.21 & \textbf{74.15} \\
      \hline
    \end{tabular}
    \caption{Results on Swin-B.}
    \label{tableswin}
  \end{minipage}
\end{table}

First, to illustrate the versatility of our GIST across models of varying sizes, we substitute ViT-B/16 with ViT-S/16 and ViT-L/16, as detailed in Table \ref{tablevit}. Next, to highlight our framework's adaptability to different network structures, we conduct experiments using Swin-B \cite{swinTRM} as the backbone, as presented in Table \ref{tableswin}. As evident from Tables \ref{tablevit} and \ref{tableswin}, regardless of whether we modify the model size or transition to an alternate backbone, our GIST consistently bolsters performance without a significant increase in parameters.

\paragraph{Comparisons with self-knowledge distillation methods}

\begin{table}[t]
\centering
\footnotesize
\setlength{\tabcolsep}{2pt}
\begin{tabular}{c|c|ccc}
\hline
Method        & Mean           & Natural        & Specialized    & Structured     \\ \hline
Adapter       & 71.46          & 79.96          & 86.02          & 56.73          \\
Adapter+BYOT  & 69.70          & 77.86          & 86.24          & 54.29          \\
Adapter+CS-KD  & 71.24          & \textbf{82.63} & 86.22          & 53.78          \\
% Adapter+Tf-KD & 72.80          & 82.26          & 86.47          & 57.70          \\ 
Adapter+USKD & 71.40         & 80.14          & 86.78          & 56.06         \\
\textbf{Adapter\textasteriskcentered}  & \textbf{73.71} & 82.26          & \textbf{87.50} & \textbf{59.24} \\ \hline
\end{tabular}
\caption{The comparative results with different self-knowledge distillation methods.}
\label{tableskd}
\end{table}

Our work is inspired by self-knowledge distillation (SKD) techniques. Consequently, we compare our approach with two classical methods (BYOT \cite{BYOT} and CS-KD \cite{CSKD}) as well as a state-of-the-art method (USKD \cite{uskd}). The results are presented in Table \ref{tableskd}, which reveal that even the most advanced SKD techniques can lead to a performance degradation of PEFT methods. A potential reason is that the existing SKD methods do not specifically acquire soft labels tailored for fine-tuning phase. In contrast to them, the Gist token we introduced serves as an aggregator, effectively capturing task-specific knowledge, thereby providing superior soft labels for knowledge interaction.

\subsection{Visualization}

\begin{figure}[t] % 图片环境，htbp参数表示浮动定位
\centering % 居中对齐
\includegraphics[width=0.45\textwidth]{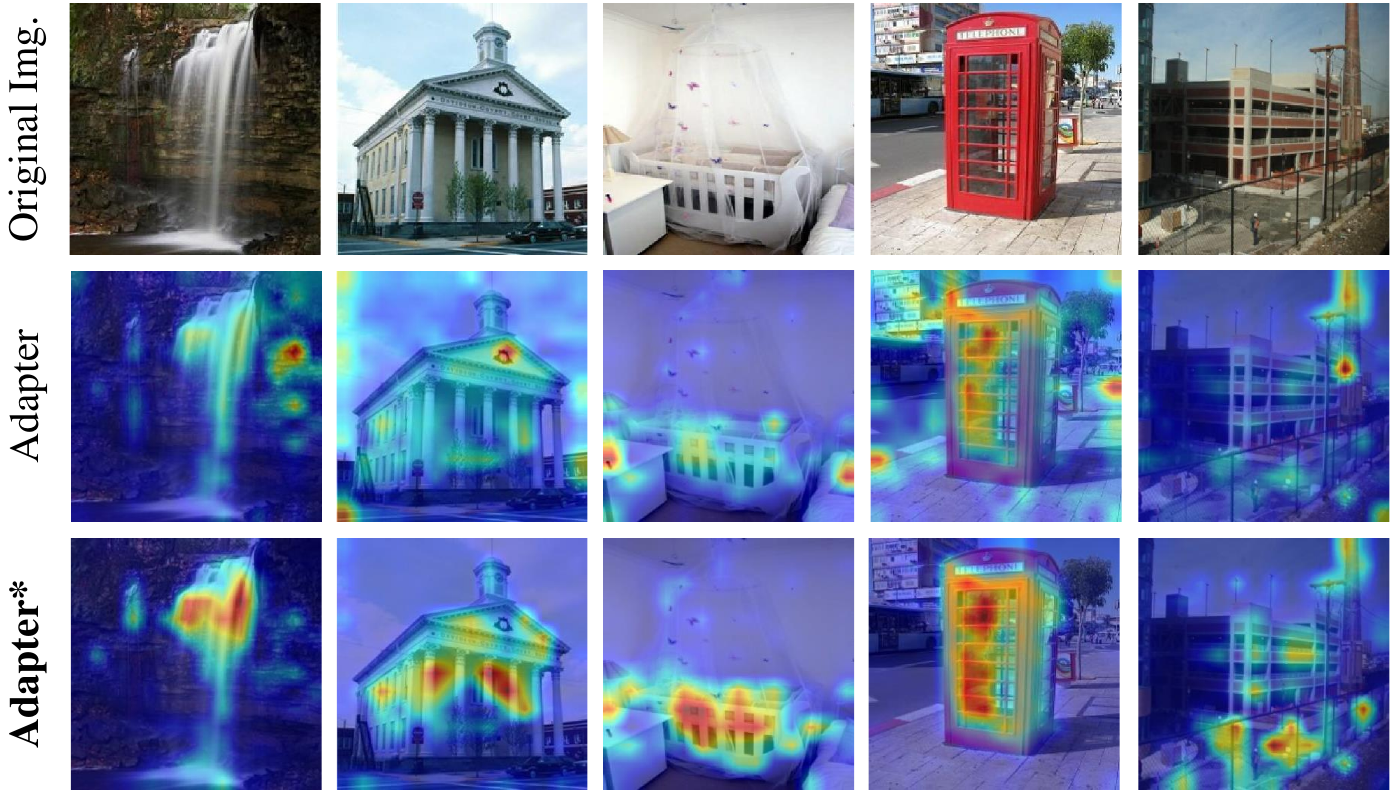} % 插入图片，width=0.5\textwidth表示图片宽度为文本宽度的一半
\caption{The attention map visualization on Sun397 dataset.} % 图片标题
\label{figureatt}
\end{figure}

\begin{figure}[t] % 图片环境，htbp参数表示浮动定位
\centering % 居中对齐
\includegraphics[width=0.45\textwidth]{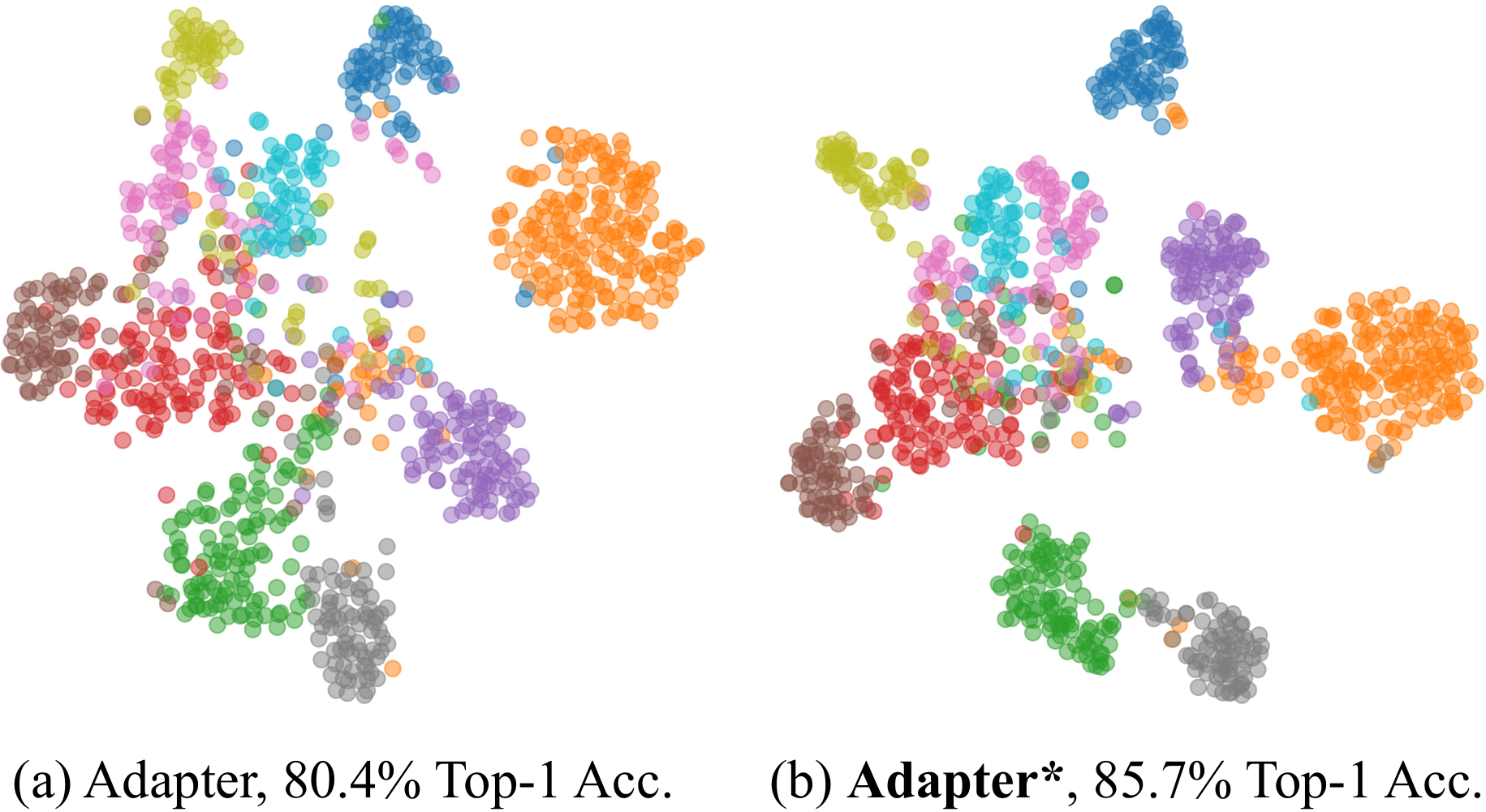} % 插入图片，width=0.5\textwidth表示图片宽度为文本宽度的一半
\caption{The t-SNE visualization on SVHN dataset.} % 图片标题
\label{figuretsne}
\end{figure}

We conduct attention map and t-SNE \cite{tsne} visualization analysis, as depicted in Figure \ref{figureatt} and \ref{figuretsne}. For this, we extract the [CLS] following the final Transformer layer and preceding the linear classification head. This analysis is performed on the Sun397 \cite{Sun397} and SVHN \cite{SVHN} dataset. Notably, upon integrating GIST, the attention is more focused on the target object, and the classification clusters appear more condensed. This suggests that our framework enhances the ability of existing PEFT methods to assimilate more thorough task-specific knowledge via knowledge interaction. More results of visualization are in the Appendix.

\subsection{Discussion}

Our GIST framework possesses the following two preeminent characteristics:

\begin{itemize}
\item \textbf{Universality}: In the experimental section, we conduct experiments for PEFT methods on the image classification, fine-grained few-shot and language understanding tasks. The results demonstrate that our framework is versatile and can be applied to PEFT methods across various scenarios, not just confined to computer vision fields.
\item \textbf{Scalability}: At the core of GIST framework lies the principle of knowledge interaction, which can be realized in multiple ways, not merely limited to the approach presented in this paper. A simple illustration, as shown in Table \ref{tabledifferentloss}, reveals that by substituting the BKLD loss with other common losses for knowledge interaction, performance can still be augmented. This means that advanced loss functions in future research can be seamlessly integrated into our GIST framework to improve the fine-tuning performance.
\end{itemize}

\section{Conclusions}

In this paper, we propose GIST, an efficient and straightforward fine-tuning framework, tailored specifically for PEFT methods. This framework incorporates a learnable Gist token to explicitly establish a connection between trainable parameters and downstream tasks, thereby aiming to acquire a more comprehensive task-specific knowledge. In addition, it employs a Bidirectional Kullback-Leibler Divergence loss to enhance the interaction between task-specific and intrinsic task-agnostic knowledge of pre-trained models. Extensive experiments demonstrate that integrating existing PEFT methods with our GIST framework leads to improved performance without significantly increasing the parameter count.
\clearpage
\setcounter{page}{1}
\maketitlesupplementary

\section{More detailed motivations}

In this paper, our primary motivation stems from the disparities between different types of knowledge. Initially, during the pre-training phase, the datasets employed are often large-scale and diverse, yet lacking specialized information. As a result, the pre-training phase mainly endows the model with task-agnostic knowledge. In contrast, the datasets used during the fine-tuning phase tend to be small-scale and specialized, embodying primarily task-specific knowledge. At this juncture, the pre-trained model is tasked with adjusting its intrinsic task-agnostic knowledge to bridge the gap with the task-specific requirements, thereby adapting to downstream tasks. However, the volume of samples in downstream datasets is significantly smaller than that of the pre-training datasets. This means that the information content of task-specific knowledge is also less than that of task-agnostic knowledge. Consequently, the Parameter-Efficient Fine-Tuning (PEFT) approach posits that it isn't necessary to update all model parameters. Instead, making adjustments to or introducing a small number of trainable parameters can suffice for acquiring task-specific knowledge during the fine-tuning phase.

However, during the fine-tuning phase, the PEFT method under the traditional framework introduces learnable parameters that lack an explicit connection with downstream targets, leading to an inadequate acquisition of downstream knowledge. To address this, we have introduced the 'Gist token', creating a bridge between the learnable parameters and downstream objectives for more effective learning of task-specific knowledge. Further enhancing this approach, we utilize knowledge interaction through a Bidirectional Kullback-Leibler Divergence loss. This method calculates the KL divergence between Class logits, representing task-agnostic knowledge, and Gist logits, which embody task-specific knowledge. Such an interaction allows for mutual guidance between these knowledge types, significantly improving the model's adaptability to downstream tasks.

\begin{figure}[!h] % 图片环境，htbp参数表示浮动定位
\centering % 居中对齐
\includegraphics[width=0.4\textwidth]{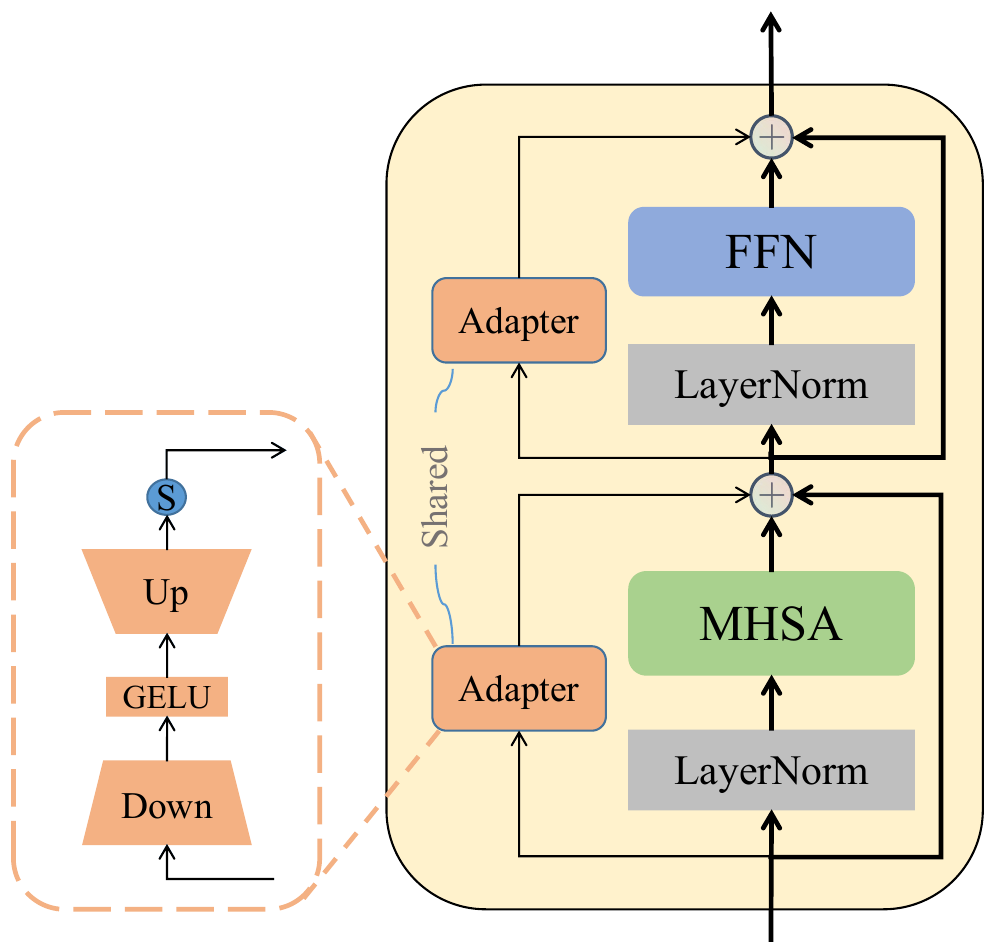}
    \caption{The Shared Adapter utilized on the VTAB-1K benchmark in this paper.}
    \label{figcvadapter}
\end{figure}

\begin{table*}[!h]
    \centering
    \begin{tabular}{c|c|ccc|c}
    \hline
    \textbf{Group} & \textbf{Dataset} &\textbf{Train} & \textbf{Val} & \textbf{Test} & \textbf{\# Class} \\ \hline
    \multirow{7}{*}{\textbf{Natural}} & \textbf{CIFAR100} \cite{cifar100} &\multirow{7}{*}{800/1,000} &\multirow{7}{*}{200} & 10,000 &100  \\
         & \textbf{Caltech101} \cite{Caltech101} & & & 6,084 &102  \\
         & \textbf{DTD} \cite{dtd} & & & 1,880 &47  \\
         & \textbf{Oxford-Flowers102} \cite{flowers102} & & & 6,149 &102  \\
         & \textbf{Oxford-Pets} \cite{OxfordPets} & & & 3,669 &37  \\
         & \textbf{SVHN} \cite{SVHN} & & & 26,032 &10  \\
         & \textbf{Sun397} \cite{Sun397} & & & 21,750 &397  \\
         \hline

    \multirow{4}{*}{\textbf{Specialized}} & \textbf{Patch Camelyon} \cite{PatchCamelyon}&\multirow{4}{*}{800/1,000} &\multirow{4}{*}{200} & 32,768 &2  \\
         & \textbf{EuroSAT} \cite{Eurosat} & & & 5,400 &10  \\
         & \textbf{Resisc45} \cite{Resisc45} & & & 6,300 &45  \\
         & \textbf{Retinopathy} \cite{Retinopathy} & & & 42,670 &5  \\
         \hline

    \multirow{8}{*}{\textbf{Structured}} & \textbf{Clevr/count} \cite{Clevr} &\multirow{8}{*}{800/1,000} &\multirow{8}{*}{200} & 15,000 &8  \\
         & \textbf{Clevr/distance} \cite{Clevr} & & & 15,000 &6  \\
         & \textbf{DMLab} \cite{DMLab} & & & 22,735 &6  \\
         & \textbf{KITTI-Dist} \cite{KITTI} & & & 711 &4  \\
         & \textbf{dSprites/location} \cite{dSprites} & & & 73,728 &16  \\
         & \textbf{dSprites/orientation} \cite{dSprites} & & & 73,728 &16  \\
         & \textbf{SmallNORB/azimuth} \cite{SmallNORB} & & & 12,150 &18  \\
         & \textbf{SmallNORB/elevation} \cite{SmallNORB} & & & 12,150 &18  \\
         \hline
    \end{tabular}
    \caption{The details of the VTAB-1K benchmark.}
    \label{tablevtab1k}
\end{table*}

\begin{table*}[!h]
    \centering
    \begin{tabular}{c|ccc|c} \hline
    \textbf{Dataset} &\textbf{Train} & \textbf{Val} & \textbf{Test} & \textbf{\# Class} \\ \hline
    \textbf{Food-101} \cite{FGVC-food101} &\multirow{5}{*}{(1/2/4/8/16)*(\#Class)} & 20,200 & 30,300 & 101 \\
    \textbf{Oxford-Pets} \cite{FGVC-pets} & & 736 & 3,669 & 37 \\
    \textbf{Stanford Cars} \cite{FGVC-cars} & & 1,635& 8,041&196 \\
    \textbf{Oxford-Flowers102} \cite{FGVC-flowers102} & &1,633&2,463&102 \\
    \textbf{FGVC-Aircraft} \cite{FGVC-aircraft} & &3,333&3,333&100 \\ \hline
    \end{tabular}
    \caption{The details of the FGVC datasets.}
    \label{tablefgvc}
\end{table*}

\begin{table*}[!h]
    \centering
    \begin{tabular}{c|ccc|cc}
    \hline
    \textbf{Dataset} & \textbf{Task} & \textbf{Domain} & \textbf{Metric}  & \textbf{Train} & \textbf{Test}\\ \hline
    \textbf{MNLI} & natural language inference & various & accuracy & 393k &20k\\
    \textbf{QQP} & paraphrase detection & social QA questions (Quora) & accuracy \& F1 & 364k &391k\\ 
    \textbf{QNLI} & natural language inference & Wikipedia & accuracy & 105k &5.4k\\ 
    \textbf{SST-2} & sentiment analysis & Movie Reviews & accuracy & 67k &1.8k\\ 
    \textbf{STS-B} & sentence similarity & various & Pearson \& Spearman corr. & 7k &1.4k\\ 
    \textbf{MRPC} & paraphrase detection & news & accuracy \& F1 & 3.7k &1.7k\\ 
    \textbf{RTE} & natural language inference & News, Wikipedia & accuracy & 2.5k &3k\\ 
    \textbf{CoLA} & acceptability & various & Matthews corr. & 8.5k &1k\\ 
         \hline
    \end{tabular}
    \caption{The details of 8 tasks we utilized on the GLUE benchmark.}
    \label{tableglue}
\end{table*}

\begin{table*}[!h]
    \centering
    \footnotesize
    \setlength{\tabcolsep}{2pt}
    \begin{tabular}{c|ccccccccccccccccccc}
    \hline
      Dataset   & \rotatebox{90}{CIFAR-100} & \rotatebox{90}{Caltech101} & \rotatebox{90}{DTD}  & \rotatebox{90}{Flowers102} & \rotatebox{90}{Pets} & \rotatebox{90}{SVHN} & \rotatebox{90}{Sun397} & \rotatebox{90}{Patch Camelyon} & \rotatebox{90}{EuroSAT} & \rotatebox{90}{Resisc45} & \rotatebox{90}{Retinopathy} & \rotatebox{90}{Clevr/count} & \rotatebox{90}{Clevr/distance} & \rotatebox{90}{DMLab} & \rotatebox{90}{KITTI/distance} & \rotatebox{90}{dSprites/loc} & \rotatebox{90}{dSprites/ori} & \rotatebox{90}{SmallNORB/azi} & \rotatebox{90}{SmallNORB/ele}  \\ \hline
      Initial lr  &5e-3 &1e-3 &5e-3 &5e-3 &5e-3 &1e-2 &5e-3 &5e-3 &3e-3 &2e-3 &5e-3 &2e-3 &5e-2 &5e-3 &1e-2 &1e-2 &5e-3 &2e-2 & 5e-3 \\
      \hline
    \end{tabular}
    \caption{The detailed initial learning rate of SSF \cite{ssf} method on the VTAB-1K benchmark.}
    \label{tablecvlr}
\end{table*}

\begin{table*}[!h]
\centering
\footnotesize
\setlength{\tabcolsep}{2pt}
\begin{tabular}{c|ccccccc|cccc|cccccccc|cc}
\hline
\multicolumn{1}{l|}{} & \multicolumn{7}{c|}{Natural}                                      & \multicolumn{4}{c|}{Specialized}                  & \multicolumn{8}{c|}{Structured}                                                                                     & \multicolumn{1}{l}{} & \multicolumn{1}{l}{} \\ \hline
Method             & \rotatebox{90}{CIFAR-100} & \rotatebox{90}{Caltech101} & \rotatebox{90}{DTD}  & \rotatebox{90}{Flowers102} & \rotatebox{90}{Pets} & \rotatebox{90}{SVHN} & \rotatebox{90}{Sun397} & \rotatebox{90}{Patch Camelyon} & \rotatebox{90}{EuroSAT} & \rotatebox{90}{Resisc45} & \rotatebox{90}{Retinopathy} & \rotatebox{90}{Clevr/count} & \rotatebox{90}{Clevr/distance} & \rotatebox{90}{DMLab} & \rotatebox{90}{KITTI/distance} & \rotatebox{90}{dSprites/loc} & \rotatebox{90}{dSprites/ori} & \rotatebox{90}{SmallNORB/azi} & \rotatebox{90}{SmallNORB/ele} & \rotatebox{90}{Mean}       & \rotatebox{90}{Params. (M)}          \\ \hline
sequential Adapter &74.1 & 86.1 &63.2 &97.7 &87.0 &34.6 &50.8 &76.3 &88.0 &73.1 &70.5 &45.7 &37.4 &31.2 &53.2 &30.3 &25.4 &13.8 &22.1  & 55.82 & 0.27 \\
Shared Adapter             & 70.2      & 92.6       & 74.6 & 99.4       & 91.2 & 80.4 & 51.4   & 84.1           & 96.3    & 88.0    & 75.6        & 84.2        & 59.6           & 53.2  & 76.3           & 60.7         & 51.9         & 27.8          & 40.2          & 71.46     & 0.13                 \\ 
\hline
\end{tabular}
\caption{The comparative results on VTAB-1K, using the backbone of ViT-B/16 pre-trained on ImageNet-21K. The results of the sequential adapter are from \cite{ssf}.}
\label{tableadapter}
\end{table*}

\section{Datasets}

In this section, we present detailed information about the VTAB-1K benchmark  \cite{vtab1k}, FGVC datasets, GLUE benchmark \cite{glue} used in this paper, as shown in Tables \ref{tablevtab1k}, \ref{tablefgvc}, and \ref{tableglue}. Notably, following the previous work \cite{attempt}, we utilize 8 tasks on the GLUE benchmark, including MNLI \cite{MNLI}, QQP \footnote{data.quora.com/First-Quora-Dataset-Release-Question-Pairs}, QNLI \cite{QNLI}, SST-2 \cite{SST-2}, STS-B \cite{STS-B}, MRPC \cite{MRPC}, RTE \cite{RTE1, RTE2, RTE3, RTE5}, and CoLA \cite{CoLA}.

\section{Implementation details on VATB-1K}

Our GIST framework is compared against the traditional fine-tuning framework. Therefore, for different PEFT methods, we keep the implementation settings the same with the traditional framework.

\subsection{Adapter, VPT and SSF}

\subsubsection{Training settings} For Adapter, VPT and SSF methods, we follow the training settings of SSF \cite{ssf}. Namely, we directly resize the image to 224 × 224. We employ AdamW \cite{adamw} as the optimizer, set the batch size to 32, and designate 100 epochs with a provision for a 10-epoch warm-up at a warmup learning rate of 1e-7. Regarding the initial learning rate (lr), previous study \cite{ssf} have established different values for various datasets, as detailed in Table \ref{tablecvlr}.

\subsubsection{Adapter method settings} 
\label{adaptersettings}

In this study, we opt for the parallel model Adapter instead of the sequential one, offering a stronger baseline (with fewer parameters and improved performance) when compared under the traditional fine-tuning framework, as illustrated in Table \ref{tableadapter}. Specifically, as depicted in Figure \ref{figcvadapter}, within the Adapter, we set the dimension of the intermediate layer to 4 and designate a scaling factor $s=0.1$ for all experiments.

\subsubsection{VPT method settings}
\label{vptsettings}
In the original paper of the VPT method \cite{VPT}, the authors conducted an extensive search on the VTAB-1K benchmark for various tasks, experimenting with the lengths of the newly introduced prompt token in the set \{1, 5, 10, 50, 100, 200\}. This search process is both tedious and intricate. However, the primary aim of our study is to contrast the advantages of the GIST framework against the traditional fine-tuning framework. Consequently, for the VPT method, we consistently set the prompt token length to 20 across all experiments.

\subsubsection{SSF method settings}
\label{ssfsettings}
SSF \cite{ssf} eliminates the need for a tedious hyperparameter search process. Therefore, we conduct experiments under our GIST framework using the default settings from the SSF source code directly.

\subsection{FacT, ReAdapter}

\subsubsection{Training settings} 
Following \cite{fact, readapter}, we resize the images to 224 × 224 and then normalize them using ImageNet's mean and standard deviation. We employ AdamW as our optimizer with a batch size set to 64. The initial learning rate is set at 1e-3, with a weight decay of 1e-4. Training is conducted over 100 epochs, inclusive of 10 warm-up epochs, and we utilize the CosineAnnealingLR \cite{cosineannealingLR} for the learning rate scheduler.

\subsubsection{FacT method settings}

In the original paper for FacT \cite{fact}, the authors undertook an extensive hyperparameter search. Specifically, for different tasks within the VTAB-1K benchmark, the authors searched for the scaling factor $s$ across \{0.01, 0.1, 1, 10, 100\}. Moreover, for the rank $r$, they searched within the set \{2, 4, 8, 16, 32\}. In our study, we directly utilized the default settings from the official FacT code to conduct experiments under the GIST framework. However, it's important to note that for the FacT method under the traditional fine-tuning framework, we directly report the results that the authors provided in the original paper, which came after their elaborate hyperparameter search. Therefore, our GIST framework operates from a potentially disadvantaged baseline. Still, the results of Table 1 demonstrate that our GIST framework manages to surpass the traditional framework by 0.32\%.

\subsubsection{ReAdapter method settings} 

In the original ReAdapter \cite{readapter} paper, the authors conducted a relatively straightforward hyperparameter search. Specifically, they only searched for the scaling factor $s$ within the set \{0.1, 0.5, 1, 5, 10\}. Consequently, for ReAdapter under our GIST framework, we carried out the same hyperparameter search. The results indicate that using ReAdapter within our GIST framework outperforms its utilization within the traditional fine-tuning framework by 0.43\%.

\begin{table*}[!h]
\centering
\footnotesize
\setlength{\tabcolsep}{2pt}
\begin{tabular}{c|ccccccc|cccc|cccccccc|c}
\hline
$\lambda$             & \rotatebox{90}{CIFAR-100} & \rotatebox{90}{Caltech101} & \rotatebox{90}{DTD}  & \rotatebox{90}{Flowers102} & \rotatebox{90}{Pets} & \rotatebox{90}{SVHN} & \rotatebox{90}{Sun397} & \rotatebox{90}{Patch Camelyon} & \rotatebox{90}{EuroSAT} & \rotatebox{90}{Resisc45} & \rotatebox{90}{Retinopathy} & \rotatebox{90}{Clevr/count} & \rotatebox{90}{Clevr/distance} & \rotatebox{90}{DMLab} & \rotatebox{90}{KITTI/distance} & \rotatebox{90}{dSprites/loc} & \rotatebox{90}{dSprites/ori} & \rotatebox{90}{SmallNORB/azi} & \rotatebox{90}{SmallNORB/ele} & \rotatebox{90}{Mean}          \\ \hline
-   & 70.2      & 92.6       & 74.6 & 99.4       & 91.2 & 80.4 & 51.4   & 84.1           & 96.3    & 88.0    & 75.6        & 84.2        & 59.6           & 53.2  & 76.3           & 60.7         & 51.9         & 27.8          & 40.2          & 71.46                \\ 
0.25 &74.3 &	92.3 	&76.9 	&99.5 	&92.0 	&85.7 &	54.5 &87.1 	&96.5 	&87.5 &	77.2 	&83.3 	&61.2 &	53.4 &	80.2 &	70.8 &	52.0 &	29.3& 	39.3  & 73.31  \\
0.5 &74.5 	&92.2 &	76.3 &	99.5 &	92.1 	&85.0 &	54.4 &	88.1 	&96.3 	&87.6 	&77.4 	&83.1 	&59.8 	&54.0 	&78.7 	&69.5 	&51.9 	&29.1 	&41.0  &73.18  \\
0.75 &74.4 &	92.3 	&76.6 	&99.5 	&92.3 	&85.3 	&54.6 	&88.2 	&96.4 	&87.9 	&76.5 	&83.6 	&60.1 	&53.0 	&81.2 	&72.3 	&52.1 	&29.2 	&39.8  & 73.44  \\
 % & & & & & & & & & & & & & & & & & & & & 
\hline
\end{tabular}
\caption{Detailed results of the ablation studies for different $\lambda$ on the VTAB-1K benchmark.}
\label{tableappendlambda}
\end{table*}

\begin{table*}[!h]
\centering
\footnotesize
\setlength{\tabcolsep}{2pt}
\begin{tabular}{c|ccccccc|cccc|cccccccc|cc}
\hline
Token length             & \rotatebox{90}{CIFAR-100} & \rotatebox{90}{Caltech101} & \rotatebox{90}{DTD}  & \rotatebox{90}{Flowers102} & \rotatebox{90}{Pets} & \rotatebox{90}{SVHN} & \rotatebox{90}{Sun397} & \rotatebox{90}{Patch Camelyon} & \rotatebox{90}{EuroSAT} & \rotatebox{90}{Resisc45} & \rotatebox{90}{Retinopathy} & \rotatebox{90}{Clevr/count} & \rotatebox{90}{Clevr/distance} & \rotatebox{90}{DMLab} & \rotatebox{90}{KITTI/distance} & \rotatebox{90}{dSprites/loc} & \rotatebox{90}{dSprites/ori} & \rotatebox{90}{SmallNORB/azi} & \rotatebox{90}{SmallNORB/ele} & \rotatebox{90}{Mean}     & \rotatebox{90}{Params. (M)}      \\ \hline
1   & 74.5     & 92.3       & 76.9 & 99.5     &92.3 & 85.7 & 54.6   & 88.2           & 96.5    & 87.9     & 77.4        & 83.6        & 61.2           & 54.0    & 81.2          & 72.3         & 52.1         & 29.3          & 41.0            & 73.71      &     0.13    \\ 
10  &74.0 &	92.3 	&76.9 	&99.6& 	92.3 	&86.3 	&54.2 	&87.2 	&96.1 	&87.4 	&75.8 	&83.6 	&61.5 	&53.3 	&79.5 	&71.3 	&52.9 	&28.2 	&42.6 &73.42&0.14\\
50  &73.0 	&92.9 	&74.9 	&99.5 	&91.9 	&87.8 	&52.9 &85.8 	&96.3 	&87.7 	&75.6 	&83.3 	&54.8 	&52.6 	&79.4 	&65.3 	&53.4 	&19.2 	&40.8  &71.96&0.16\\
100  &72.9 	&92.9 	&73.1 	&99.4 	&91.1 	&88.1 	&52.2 	&85.3 &	96.3 &	86.9 	&76.8 	&81.7 	&37.2 &	52.0 	&79.9 	&66.1 &	52.2 	&28.8 	&40.2  &71.21&0.21\\
 % & & & & & & & & & & & & & & & & & & & & 
\hline
\end{tabular}
\caption{Detailed results of the ablation studies for different Gist token length on the VTAB-1K benchmark.}
\label{tableappendlen}
\end{table*}

\begin{table*}[!h]
\centering
\footnotesize
\setlength{\tabcolsep}{2pt}
\begin{tabular}{cccc|ccccccc|cccc|cccccccc|c}
\hline
$\mathcal{L}_{cls}$   & $\mathcal{L}_{gist}$   &  $\mathcal{L}_{fkl}$    & $\mathcal{L}_{rkl}$               & \rotatebox{90}{CIFAR-100} & \rotatebox{90}{Caltech101} & \rotatebox{90}{DTD}  & \rotatebox{90}{Flowers102} & \rotatebox{90}{Pets} & \rotatebox{90}{SVHN} & \rotatebox{90}{Sun397} & \rotatebox{90}{Patch Camelyon} & \rotatebox{90}{EuroSAT} & \rotatebox{90}{Resisc45} & \rotatebox{90}{Retinopathy} & \rotatebox{90}{Clevr/count} & \rotatebox{90}{Clevr/distance} & \rotatebox{90}{DMLab} & \rotatebox{90}{KITTI/distance} & \rotatebox{90}{dSprites/loc} & \rotatebox{90}{dSprites/ori} & \rotatebox{90}{SmallNORB/azi} & \rotatebox{90}{SmallNORB/ele} & \rotatebox{90}{Mean}          \\ \hline
\checkmark &  &     &         & 70.2      & 92.6       & 74.6 & 99.4       & 91.2 & 80.4 & 51.4   & 84.1           & 96.3    & 88.0    & 75.6        & 84.2        & 59.6           & 53.2  & 76.3           & 60.7         & 51.9         & 27.8          & 40.2          & 71.46                \\ 
\checkmark & \checkmark & \checkmark & \checkmark  & 74.5     & 92.3       & 76.9 & 99.5     &92.3 & 85.7 & 54.6   & 88.2           & 96.5    & 87.9     & 77.4        & 83.6        & 61.2           & 54.0    & 81.2          & 72.3         & 52.1         & 29.3          & 41.0            & 73.71    \\ 

\hline

\checkmark &            & \checkmark & \checkmark &73.9 	&91.9 	&76.6 	&99.5 	&92.2 	&85.4 	&54.2 	&87.6 	&96.4 	&88.2 	&76.6 	&83.7 	&60.2 	&53.6 	&79.9 	&70.4 	&53.2 	&28.4 	&40.3  &73.29  \\

\checkmark & \checkmark &             & \checkmark &74.3 	&92.3 	&76.6 	&99.5 &	92.1 &	85.0 	&54.6 	&87.8 	&96.7 	&87.7 	&77.1 	&83.5 	&61.5 	&53.5 	&80.5 	&70.8 	&52.9 	&29.6 	&40.2   & 73.48  \\

\checkmark & \checkmark & \checkmark & &74.3 	&92.2 	&76.8 	&99.5 	&92.1 	&85.4 	&54.6 	&87.5 	&96.6 	&87.8 	&76.2 	&83.5 	&60.5 	&53.7 	&79.8 	&68.2 	&51.8 	&27.0 	&41.0   & 73.07  \\ 

\hline

 \checkmark &     &                & \checkmark &73.4 	&91.9 	&76.5 	&99.5 	&92.1 	&84.8 &	54.0 	&87.4 	&96.7 	&87.5 	&76.8 &	83.7 &	60.5& 	53.2& 	79.2 &	66.4 &	52.7 &	26.5 	&39.7   & 72.76  \\

 \checkmark &            & \checkmark & &72.5 &92.0 	&75.9 	&99.5 	&92.2 &	83.0 	&53.8 	&84.2 	&96.4 	&87.6 	&76.6 	&84.0 &59.8 	&53.6 	&78.2 &69.0 	&52.4 	&27.0 &	39.3   & 72.46  \\

\checkmark & \checkmark &       & &72.2 	&92.3 	&75.3 	&99.4 &	91.8 	&83.0 	&53.7 &86.4 	&96.5 &	87.6 &	76.7 &	83.6 	&61.0 	&53.1 	&79.1 	&67.0 	&52.1 	&28.9 	&41.9   & 72.71  \\
 % & & & & & & & & & & & & & & & & & & & & 
\hline
\end{tabular}
\caption{Detailed results of the ablation studies of our loss function on the VTAB-1K benchmark.}
\label{tableappendourloss}
\end{table*}

\begin{table*}[!h]
\centering
\footnotesize
\setlength{\tabcolsep}{2pt}
\begin{tabular}{l|ccccccc|cccc|cccccccc|c}
\hline
Loss function             & \rotatebox{90}{CIFAR-100} & \rotatebox{90}{Caltech101} & \rotatebox{90}{DTD}  & \rotatebox{90}{Flowers102} & \rotatebox{90}{Pets} & \rotatebox{90}{SVHN} & \rotatebox{90}{Sun397} & \rotatebox{90}{Patch Camelyon} & \rotatebox{90}{EuroSAT} & \rotatebox{90}{Resisc45} & \rotatebox{90}{Retinopathy} & \rotatebox{90}{Clevr/count} & \rotatebox{90}{Clevr/distance} & \rotatebox{90}{DMLab} & \rotatebox{90}{KITTI/distance} & \rotatebox{90}{dSprites/loc} & \rotatebox{90}{dSprites/ori} & \rotatebox{90}{SmallNORB/azi} & \rotatebox{90}{SmallNORB/ele} & \rotatebox{90}{Mean}          \\ \hline
$\mathcal{L}_{cls}$   & 70.2      & 92.6       & 74.6 & 99.4       & 91.2 & 80.4 & 51.4   & 84.1           & 96.3    & 88.0    & 75.6        & 84.2        & 59.6           & 53.2  & 76.3           & 60.7         & 51.9         & 27.8          & 40.2          & 71.46                \\ 
$\mathcal{L}_{cls}$+$\mathcal{L}_{gist}$+$\mathcal{L}_{mse}$ &74.6	&92.6	&76.5	&99.6	&92.2	&87.1	&53.2	&88.1	&96.6	&87.6	&75.9	&81.7	&61.3	&52.0	&82.6	&65.1	&52.9	&28.8	&39.5  & 73.03  \\
$\mathcal{L}_{cls}$+$\mathcal{L}_{gist}$+$\mathcal{L}_{cos}$ &73.0	&92.0	&75.8	&99.5	&92.2	&82.9	&53.6	&86.2	&96.4	&87.5	&76.5	&83.6	&61.3	&53.7	&80.1	&69.1	&52.6	&28.0	&40.7  &72.88  \\
$\mathcal{L}_{cls}$+$\mathcal{L}_{gist}$+$\mathcal{L}_{bkl}$ & 74.5     & 92.3       & 76.9 & 99.5     &92.3 & 85.7 & 54.6   & 88.2           & 96.5    & 87.9     & 77.4        & 83.6        & 61.2           & 54.0    & 81.2          & 72.3         & 52.1         & 29.3          & 41.0            & 73.71  \\
 % & & & & & & & & & & & & & & & & & & & & 
\hline
\end{tabular}
\caption{Detailed results on different loss functions for knowledge interaction on the VTAB-1K benchmark.}
\label{tableappendloss}
\end{table*}

\begin{table*}[!h]
\centering
\footnotesize
\setlength{\tabcolsep}{2pt}
\begin{tabular}{c|ccccccc|cccc|cccccccc|cc}
\hline
Method & \rotatebox{90}{CIFAR-100} & \rotatebox{90}{Caltech101} & \rotatebox{90}{DTD}  & \rotatebox{90}{Flowers102} & \rotatebox{90}{Pets} & \rotatebox{90}{SVHN} & \rotatebox{90}{Sun397} & \rotatebox{90}{Patch Camelyon} & \rotatebox{90}{EuroSAT} & \rotatebox{90}{Resisc45} & \rotatebox{90}{Retinopathy} & \rotatebox{90}{Clevr/count} & \rotatebox{90}{Clevr/distance} & \rotatebox{90}{DMLab} & \rotatebox{90}{KITTI/distance} & \rotatebox{90}{dSprites/loc} & \rotatebox{90}{dSprites/ori} & \rotatebox{90}{SmallNORB/azi} & \rotatebox{90}{SmallNORB/ele} & \rotatebox{90}{Mean}   & \rotatebox{90}{Params. (M)}        \\ \hline
S+Adapter     &68.1 	&92.3 	&73.1 	&99.4 	&90.3 	&81.9 	&52.2 	&87.3 	&96.1 	&85.8 	&77.0 	&81.9 	&60.3 	&49.7 	&75.6 	&67.2 	&50.4 	&25.5 	&42.2 	&71.39 &0.07     \\ 
\textbf{S+Adapter\textasteriskcentered}   &70.4 	&92.4 	&74.2 	&99.4 	&90.9 &	85.2 &	52.2 	&86.4 	&96.2 	&85.6 	&76.5 	&83.1 	&61.2 	&52.3 	&77.7 	&70.4 	&50.8 	&28.2 	&43.8 	&72.47  &0.07\\
\hline
L+Adapter  &72.8 	&91.8 	&74.4 	&99.5 	&92.2 	&84.0 	&54.0 	&87.1 &	96.2 	&89.1 	&75.6 	&78.6 &57.0 	&52.6 	&77.5 	&68.2 	&53.4 	&25.7 	&34.8 	&71.81  &0.30 \\
\textbf{L+Adapter\textasteriskcentered}  &77.3 	&91.7 	&77.5 	&99.6 	&92.9 &	88.2 &	58.5 	&87.5 	&96.6 	&89.8 &	76.4 &	81.5 &	55.7 	&54.8 	&81.9 	&73.7 &	54.1	&27.6 	&38.5 	&73.89  &0.30 \\
 % & & & & & & & & & & & & & & & & & & & & 
\hline
\end{tabular}
\caption{Detailed results for ViT-S/16 (S) and ViT-L/16 (L) \cite{TRM} on the VTAB-1K benchmark. The symbol \textbf{\textasteriskcentered} indicates employing the PEFT method within our GIST framework.}
\label{tableappendvit}
\end{table*}

\begin{table*}[!h]
\centering
\footnotesize
\setlength{\tabcolsep}{2pt}
\begin{tabular}{c|ccccccc|cccc|cccccccc|cc}
\hline
Method & \rotatebox{90}{CIFAR-100} & \rotatebox{90}{Caltech101} & \rotatebox{90}{DTD}  & \rotatebox{90}{Flowers102} & \rotatebox{90}{Pets} & \rotatebox{90}{SVHN} & \rotatebox{90}{Sun397} & \rotatebox{90}{Patch Camelyon} & \rotatebox{90}{EuroSAT} & \rotatebox{90}{Resisc45} & \rotatebox{90}{Retinopathy} & \rotatebox{90}{Clevr/count} & \rotatebox{90}{Clevr/distance} & \rotatebox{90}{DMLab} & \rotatebox{90}{KITTI/distance} & \rotatebox{90}{dSprites/loc} & \rotatebox{90}{dSprites/ori} & \rotatebox{90}{SmallNORB/azi} & \rotatebox{90}{SmallNORB/ele} & \rotatebox{90}{Mean}   & \rotatebox{90}{Params. (M)}        \\ \hline
Adapter  &70.2 	&93.3 	&77.3 	&99.6 &	92.4 	&82.1 	&54.9 	&87.9 	&96.1 	&88.3& 	76.8 	&84.6 &	56.2 &	52.8 	&83.6 	&78.2 	&54.2 	&24.4 	&37.9 	&73.19   &0.21 \\
\textbf{Adapter\textasteriskcentered}  &71.6 	&93.5 	&77.9 &	99.6 &	92.6 &	85.4 &	55.6 &	88.9 	&96.7 	&88.7 	&77.0 	&84.6 	&60.4 	&54.3 	&85.3 	&78.9 	&53.1 	&26.8 	&38.0 &	74.15   &0.21 \\
 % & & & & & & & & & & & & & & & & & & & & 
\hline
\end{tabular}
\caption{Detailed results for Swin-B \cite{swinTRM} on the VTAB-1K benchmark. The symbol \textbf{\textasteriskcentered} indicates employing the PEFT method within our GIST framework.}
\label{tableappendswin}
\end{table*}

\begin{table*}[!h]
\centering
\footnotesize
\setlength{\tabcolsep}{2pt}
\begin{tabular}{c|ccccccc|cccc|cccccccc|c}
\hline
Method            & \rotatebox{90}{CIFAR-100} & \rotatebox{90}{Caltech101} & \rotatebox{90}{DTD}  & \rotatebox{90}{Flowers102} & \rotatebox{90}{Pets} & \rotatebox{90}{SVHN} & \rotatebox{90}{Sun397} & \rotatebox{90}{Patch Camelyon} & \rotatebox{90}{EuroSAT} & \rotatebox{90}{Resisc45} & \rotatebox{90}{Retinopathy} & \rotatebox{90}{Clevr/count} & \rotatebox{90}{Clevr/distance} & \rotatebox{90}{DMLab} & \rotatebox{90}{KITTI/distance} & \rotatebox{90}{dSprites/loc} & \rotatebox{90}{dSprites/ori} & \rotatebox{90}{SmallNORB/azi} & \rotatebox{90}{SmallNORB/ele} & \rotatebox{90}{Mean}          \\ \hline
Adapter   & 70.2      & 92.6       & 74.6 & 99.4       & 91.2 & 80.4 & 51.4   & 84.1           & 96.3    & 88.0    & 75.6        & 84.2        & 59.6           & 53.2  & 76.3           & 60.7         & 51.9         & 27.8          & 40.2          & 71.46                \\ 
Adapter+BYOT &62.8 	&92.2 	&71.3 	&99.2 &	89.2 &	83.4 &	46.8 	&85.2 	&96.3 	&86.9 &	76.5 	&81.8 	&49.7 	&52.9 	&59.0 	&53.8 	&76.5 	&23.2 	&37.4 	&69.70   \\
Adapter+CS-KD &77.6 	&94.0 	&75.2 	&99.7 	&91.7 	&88.3 	&51.8 &	83.5& 	96.7 &	88.3 	&76.4 	&80.1 	&28.1 &	51.7 	&74.4 	&52.9 	&79.5 	&24.9 &	38.7 	&71.24   \\
Adapter+USKD &73.0 	&92.2 	&72.2 	&99.5 	&91.4 	&80.5 	&52.1 	&85.9 &	96.7 	&88.0 	&76.6 &	83.4 	&58.9 	&53.6 	&57.6 	&53.5 	&77.9 &	26.2 	&37.4 	&71.40   \\
\textbf{Adapter\textasteriskcentered}  & 74.5     & 92.3       & 76.9 & 99.5     &92.3 & 85.7 & 54.6   & 88.2           & 96.5    & 87.9     & 77.4        & 83.6        & 61.2           & 54.0    & 81.2          & 72.3         & 52.1         & 29.3          & 41.0            & 73.71    \\ 
 % & & & & & & & & & & & & & & & & & & & & 
\hline
\end{tabular}
\caption{Detailed results with different self-knowledge distillation methods on the VTAB-1K benchmark. The symbol \textbf{\textasteriskcentered} indicates employing the PEFT method within our GIST framework.}
\label{tableappendskd}
\end{table*}

\begin{figure*}[!h]
    \centering
\includegraphics[width=0.8\textwidth, height=0.99\textwidth]{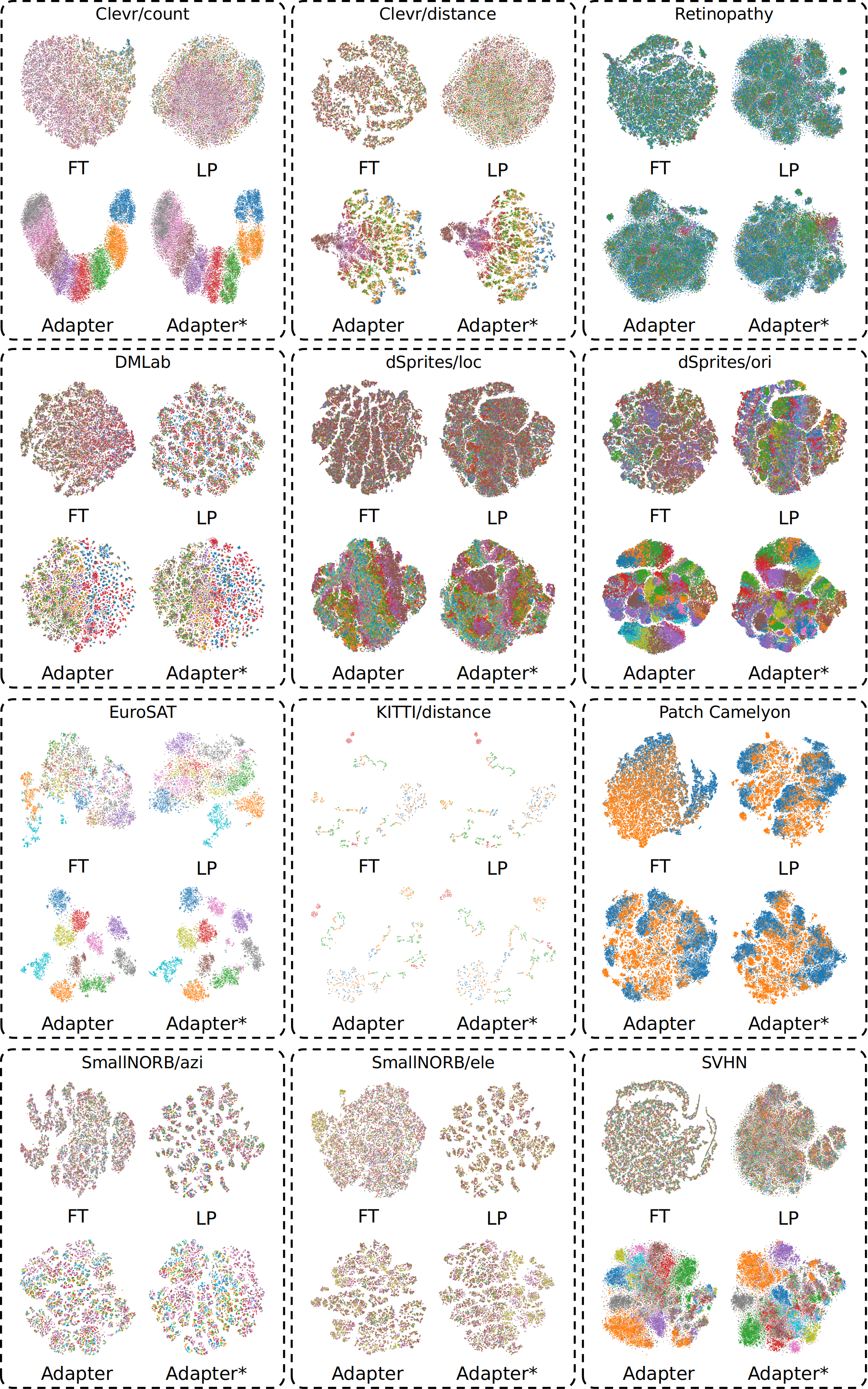}
    \caption{The results of visualization. We selected the datasets from the VTAB-1K benchmark with fewer than 20 categories for visualization. FT stands for full parameter fine-tuning, and LP stands for Linear Probing.  The symbol \textbf{\textasteriskcentered} indicates employing the PEFT method within our GIST framework.}
    \label{figureappvis}
\end{figure*}

\section{Implementation details on FGVC datasets}
For the FGVC datasets, we use Adapter, VPT \cite{VPT}, and SSF \cite{ssf} as representatives of three different PEFT methods, and have verified them in \{1, 2, 4, 8, 16\}-shot scenarios respectively.

\subsection{Training settings} For the three different PEFT methods, we consistently use AdamW \cite{adamw} as the optimizer, with a batch size set to 64, a learning rate of 1e-3, weight decay of 1e-3, training for 100 epochs with 10 warmup-epochs.

\subsection{Methods settings} For the three PEFT methods, the setup is the same as in Sec. \ref{adaptersettings}, \ref{vptsettings}, \ref{ssfsettings}.

\section{Implementation details on GLUE benchmark}

\subsection{Training settings} For the eight tasks within the GLUE benchmark, we employ consistent training configurations \cite{attempt}. Specifically, we set the batch size to 32, the max token length to 256, and the learning rate to 3e-4. Training was conducted over 20 epochs, incorporating 500 warm-up iterations.

\subsection{Method settings} For NLP tasks on the GLUE benchmark, we carry out relatively straightforward experiments to further demonstrate the universality of our GIST framework. We employ the default parameters from Adapter\cite{compacter, attempt} for our experiments. Specifically, we use GELU \cite{gelu} as the activation function and set the reduction factor to 32.

\section{More experimental results}

Due to space constraints, when conducting ablation experiments on the VTAB-1K benchmark, we only present the arithmetic mean of the Top-1 accuracy. Therefore, we display the complete results of all experiments, as shown in Tables \ref{tableappendlambda}, \ref{tableappendloss}, \ref{tableappendlen}, \ref{tableappendourloss}, \ref{tableappendvit}, \ref{tableappendswin}, and \ref{tableappendskd}.

\section{Visualization}

Due to space constraints in the main paper. Thus, we present the more visualization results for the VTAB-1K benchmark, as depicted in Figure \ref{figureappvis}.

{
    \small
    \bibliographystyle{ieeenat_fullname}
    \bibliography{main}
}


\end{document}